\documentclass{article}
\usepackage{iclr2027_conference,times}
\ifdefined\EvoSceneAnonymous
\else
  \iclrfinalcopy
\fi
\let\cite\citep

\usepackage[utf8]{inputenc} %
\usepackage[T1]{fontenc}    %
\usepackage{hyperref}       %
\usepackage{url}            %
\usepackage{booktabs}       %
\usepackage{amsfonts}       %
\usepackage{amssymb}        %
\usepackage{nicefrac}       %
\usepackage{microtype}      %
\usepackage{xcolor}         %
\usepackage{colortbl}       %
\definecolor{myblue}{RGB}{238,238,238}
\newcolumntype{Y}{>{\columncolor{myblue}\centering\arraybackslash}p{2.4em}}
\usepackage{amsmath}
\usepackage[capitalise,nameinlink]{cleveref}
\crefname{equation}{Eq.}{Eqs.}
\Crefname{equation}{Eq.}{Eqs.}
\crefname{section}{Sec.}{Secs.}
\Crefname{section}{Sec.}{Secs.}
\crefname{subsection}{Sec.}{Secs.}
\Crefname{subsection}{Sec.}{Secs.}
\crefname{subsubsection}{Sec.}{Secs.}
\Crefname{subsubsection}{Sec.}{Secs.}
\crefname{table}{Tab.}{Tabs.}
\Crefname{table}{Tab.}{Tabs.}
\crefname{figure}{Fig.}{Figs.}
\Crefname{figure}{Fig.}{Figs.}
\crefname{appendix}{Appendix}{Appendices}
\Crefname{appendix}{Appendix}{Appendices}
\usepackage{graphicx}
\usepackage[normalem]{ulem}
\usepackage{graphicx}
\usepackage{booktabs}
\usepackage{multirow}
\usepackage{makecell}
\usepackage{algorithm}
\usepackage{algpseudocode}
\usepackage{caption}
\usepackage{float}
\usepackage{placeins}
\usepackage{pifont}
\hypersetup{pdftitle={EvoScene-VLA: Evolving Scene Beliefs Inside the Action Decoder for Chunked Robot Control}}
\ificlrfinal
  \hypersetup{pdfauthor={Chushan Zhang, Ruihan Lu, Jinguang Tong, Xuesong Li, Yikai Wang, Hongdong Li}}
\else
  \hypersetup{pdfauthor={}}
\fi
\newcommand{\mkstart}{\raisebox{-0.1ex}{\scalebox{0.62}{\ding{108}}}}
\newcommand{\mkend}{\raisebox{-0.15ex}{\scalebox{0.78}{\ding{56}}}}
\newcommand{\mkok}{\raisebox{-0.15ex}{\scalebox{0.78}{\ding{52}}}}

\newcommand{\revadd}[1]{#1}
\newcommand{\revfit}[1]{#1}
\newcommand{\locnew}[1]{#1}

\title{EvoScene-VLA: Evolving Scene Beliefs\\
Inside the Action Decoder\\
for Chunked Robot Control}

\author{%
  \begin{minipage}{\dimexpr\textwidth-2\tabcolsep\relax}
    \centering
    Chushan Zhang\textsuperscript{\scriptsize 1}\hspace{0.25em}%
    Ruihan Lu\textsuperscript{\scriptsize 2}\hspace{0.25em}%
    Jinguang Tong\textsuperscript{\scriptsize 1}\hspace{0.25em}%
    Xuesong Li\textsuperscript{\scriptsize 1}\hspace{0.25em}%
    Yikai Wang\textsuperscript{\scriptsize 3,*}\hspace{0.25em}%
    Hongdong Li\textsuperscript{\scriptsize 1,*}\\[0.6ex]
    {\normalfont\small
    \textsuperscript{1}Australian National University\hspace{1.2em}%
    \textsuperscript{2}The University of Queensland\hspace{1.2em}%
    \textsuperscript{3}Beijing Normal University}
  \end{minipage}%
}

\begin{document}

\maketitle
\suppressfloats[t]
\ificlrfinal
  \lhead{Preprint}
  \begingroup
    \renewcommand{\thefootnote}{\fnsymbol{footnote}}
    \footnotetext[1]{Corresponding authors.}
  \endgroup
\else
  \lhead{Under review as a conference paper at ICLR 2027}
\fi

\begin{abstract}
Chunked vision–language–action (VLA) policies generate multi-step actions from one observation and typically re-observe after executing the action chunk.
Earlier actions change the scene conditions for later steps, while occlusion can compromise visual feedback.
The previous scene estimate is typically not propagated to the next control call.
Although spatial and temporal VLAs enhance geometric modeling and maintain temporal memory, respectively, their internal representations may become stale as robot actions alter object poses and contact states.
We introduce EvoScene-VLA, which uses compact scene tokens to unify within-chunk scene prediction, cross-chunk state propagation, and observation-based correction.
The action expert jointly denoises actions and future scene states in a single flow-matching process, allowing predicted scene changes to inform action generation.
The final scene state serves as the next prior for subsequent correction.
At the next control call, the vision–language model (VLM) fuses the incoming visual input with this prior.
To preserve control-relevant geometry, two-level geometric anchoring applies local depth supervision to per-view scene features and global 3D feature supervision to the fused scene state; a shared geometric decoder imposes the same constraints on current and future states.
Since demonstrations do not directly provide future scene states, we train a scene predictor using 3D teacher features from future images.
Its predictions provide targets for the action expert's future scene states.
At deployment, we remove both Scene Predictor and Geometric Anchor.
EvoScene-VLA improves average success over the matched baseline by 2.4 and 2.2 percentage points on 31 RoboTwin tasks under fixed and randomized settings, respectively, and by 2.7 points on LIBERO.
Across three real-robot tasks, average success rises from 37.3\% to 42.0\%.
Ablations show that a recurrent scene state alone does not necessarily improve performance, whereas state propagation can improve control when combined with geometric and future scene supervision.
\end{abstract}

\section{Introduction}

\begin{figure}[t]
\centering
\includegraphics[width=\textwidth]{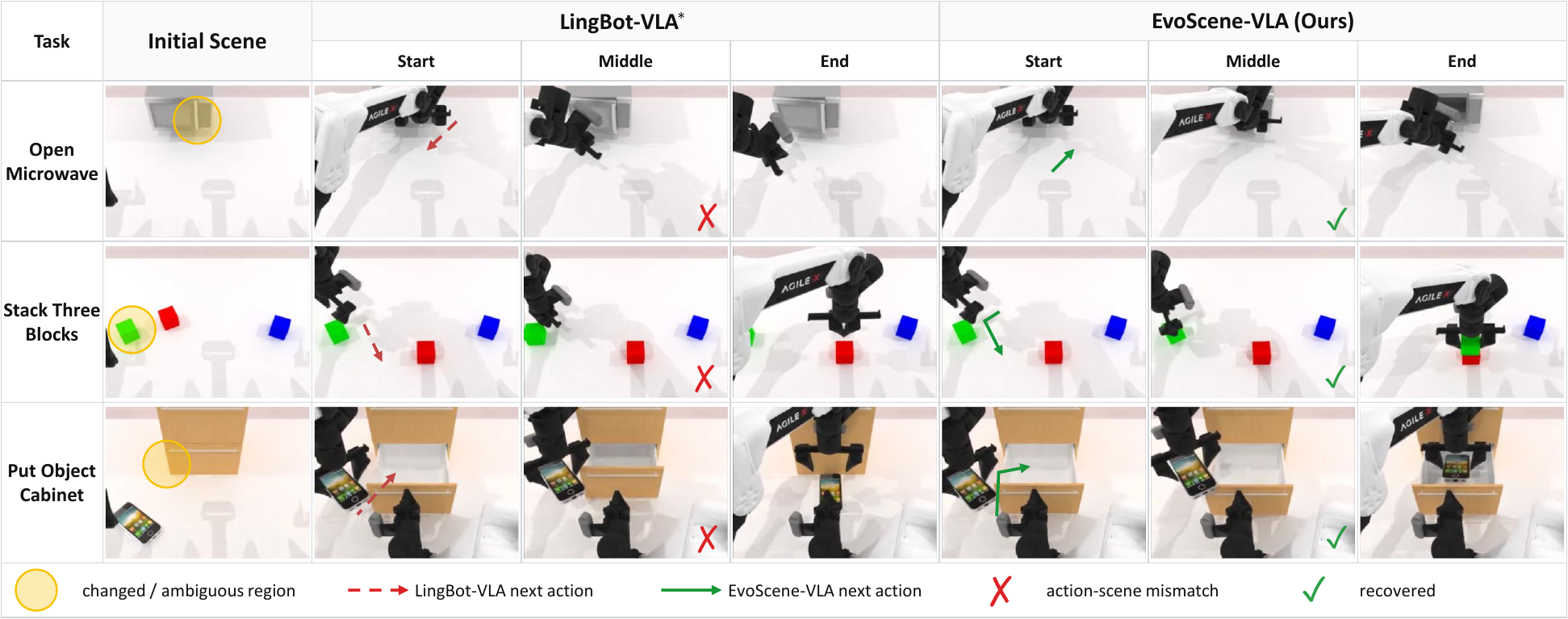}
\vspace{-1mm}
\caption{
\textbf{Qualitative analysis of chunked-control failures.}
During an action chunk, the scene can change after the planning observation is captured.
In the three RoboTwin examples shown (microwave handle, block stacking, cabinet placement), the baseline stalls mid-chunk (\textcolor{red}{\mkend}), whereas EvoScene-VLA completes the rollouts (\textcolor{green!50!black}{\mkok}).
EvoScene-VLA carries an action-updated scene prior into each chunk and co-denoises actions with predicted scene states at selected steps.}
\label{fig:qualitative_analysis_a}
\vspace{-1em}
\end{figure}

Robots change their surroundings through actions and adjust subsequent motions based on the resulting scene.
For example, to place an object into a drawer, a robot may need to adjust its trajectory to the drawer's updated position after pulling it open (\cref{fig:qualitative_analysis_a}).
However, chunked vision–language–action (VLA) policies generate multi-step action sequences from a single observation.
Visual feedback is typically acquired after execution of the action chunk.
Within a chunk, later actions remain conditioned on the initial observation despite scene changes caused by earlier actions.
This mismatch may grow as the chunk proceeds.
Furthermore, between chunks, occlusion may leave task-relevant scene geometry only partially visible in new observations~\cite{bridgevlapp}.
The previous scene estimate is typically not propagated across control calls, which can lead to spatial errors or repeated execution of completed steps.
Spatial and temporal VLAs improve scene representations derived from observations, but these representations are rarely updated to reflect the effects of the generated actions.
Spatial VLAs~\cite{spatialvla,qdepthvla,spatialforcing,3dsvla,vla4d} model geometric relationships through depth supervision, 3D encoding, or multi-view features, but these representations describe the scene before action execution.
Temporal VLAs~\cite{memoryvla,hamlet,hifvla,embodiedslotssm,tracevla} retain past observations, which can provide information when objects are temporarily occluded; after an object is moved, however, these observations may still reflect its previous position.
Action-conditioned prediction methods~\cite{flare,upvla} estimate future scene representations, but using these predictions both for action generation within a chunk and for state propagation across chunks remains underexplored.

We propose EvoScene-VLA to address the scene mismatch within a chunk and partial observations across chunks.
Within a chunk, the model jointly predicts actions and future scene states, allowing later actions to account for scene changes caused by earlier actions.
Specifically, we introduce a compact set of scene tokens to represent the scene state.
The action expert jointly denoises the action sequence and future scene states at sparse key time steps in a single flow-matching process, with the generation of the current action chunk conditioned on the future scene representations.
Across chunks, the scene state predicted at the end of an action chunk corresponds to the time of the next observation.
We therefore pass this prediction to the next control call as a scene prior.
The vision–language model (VLM) fuses the current visual observation with the scene prior, supplementing incomplete visual evidence under occlusion.
Because prediction errors and execution deviations may cause the scene prior to differ from the actual scene, the VLM also corrects the prior with the new observation during fusion.
Within-chunk prediction, cross-chunk propagation, and observation-based correction thus form a single scene-state loop, in which the scene state guides action generation in the current call and serves as a prior for correction in the next call.

However, the scene state also needs to preserve the geometric relationships needed for control.
Adding scene tokens alone does not ensure that these relationships are preserved.
To this end, we introduce a two-level \textbf{Geometric Anchor}.
\textit{Local} depth supervision constrains per-view scene features, while \textit{global} feature supervision from a 3D foundation model (3DFM) constrains the fused scene state.
We use a shared geometric decoder to align current and predicted future scene states with 3DFM teacher features.
This design provides explicit learning objectives for state prediction and propagation.

Furthermore, we train the action expert to generate future scene states using compact state targets, which are not directly available in demonstration data.
We therefore introduce a training-only \textbf{Scene Predictor}.
It predicts future scene states from the current scene state and demonstration actions.
We supervise these predictions with 3DFM teacher features extracted from future images.
These predicted scene states serve as supervision targets for the action expert's scene flow-matching objective.
At deployment, we remove both Scene Predictor and Geometric Anchor, retaining only the recurrent scene state and the joint generation of actions and scene states.

On 31 RoboTwin tasks~\cite{robotwin}, EvoScene-VLA raises the matched baseline's average success rate from 86.4\% to 88.8\% under fixed evaluation and from 85.7\% to 87.9\% under randomized initial conditions.
The comparable gains across the two settings, together with the recurrence ablations (\cref{tab:ablations}), suggest that the scene prior remains effective under pose and layout variation.
On LIBERO~\cite{liu2023libero}, EvoScene-VLA improves average success over the matched baseline by 2.7 percentage points.
On the Galaxea R1-Lite~\cite{galaxea_openworld} dual-arm platform, EvoScene-VLA achieves the highest observed success rate on all three cleaning tasks, raising average success from 37.3\% to 42.0\%.
Ablations further show that a recurrent scene state alone does not necessarily improve performance, whereas state propagation can improve control when combined with geometric and future scene supervision.
Prior interventions also show higher closed-loop success with matched than mismatched priors (\cref{tab:prior_content}),
supporting the contribution of the prior's content to control.

Our main contributions are as follows:
\begin{itemize}
\item We introduce a recurrent scene interface bridging within-chunk scene prediction and cross-chunk state updates.
Predicted future scene states condition current action generation, while the final state is carried forward as a prior and corrected using new observations.
\item We formulate the action expert with future scene-state supervision and joint flow matching to generate actions and future scene states simultaneously, eliminating the need for an additional scene predictor at deployment.
\item We develop local depth supervision with global 3D feature supervision and use a shared geometric decoder to align current and predicted future scene states with 3DFM teacher features, providing a common geometric basis for learning scene states.
\end{itemize}

\section{Related Work}

\revfit{VLA policies use VLM backbones to map images and instructions to chunked motor controls.} RT-2~\cite{rt2}, $\pi_0$~\cite{pi0}, and OpenVLA~\cite{openvla} establish this paradigm. \revfit{EvoScene-VLA adds a recurrent scene prefix to this decoder. We review geometric, temporal, and flow-based foundations.}
\paragraph{Geometric scene representation.}
Spatial VLAs add geometry to the VLM prefix or visual encoder. SpatialVLA~\cite{spatialvla} projects 3D-aware features into a spatial token grid, QDepth-VLA~\cite{qdepthvla} supervises quantized depth tokens, Spatial Forcing~\cite{spatialforcing} enforces geometric consistency, and 3DS-VLA~\cite{3dsvla} and VLA-4D~\cite{vla4d} add multi-view or temporal 3D features. \revfit{A parallel line uses 3D scene state:} PerAct~\cite{peract}, RVT~\cite{rvt,rvt2}, 3D Diffusion Policy~\cite{dp3}, and ManiGaussian~\cite{manigaussian}. DUSt3R~\cite{dust3r} and MASt3R~\cite{mast3r} recover 3D geometry from images. CUT3R~\cite{cut3r} maintains a persistent scene state that is updated with new observations. EvoScene-VLA advances a compact geometric prior with the generated action chunk. The next VLM call corrects this prior using the new observation.

\paragraph{Temporal memory and action-conditioned prediction.}
\revfit{Temporal VLAs retain observed history across calls.} MemoryVLA~\cite{memoryvla} \revfit{keeps} a summary-token memory, HAMLET~\cite{hamlet} and HiF-VLA~\cite{hifvla} use history-aware attention, Embodied-SlotSSM~\cite{embodiedslotssm} applies slot state-space models, TraceVLA~\cite{tracevla} prompts the VLM with visual traces, and AVA-VLA~\cite{avavla} aligns action and visual history. Related approaches learn trajectory representations from video~\cite{atm,vpp} or transfer across heterogeneous demonstrations~\cite{hpt}. Other methods predict \revadd{future states} from actions. FLARE~\cite{flare} inserts learnable future tokens into the VLM, UP-VLA~\cite{upvla} \revfit{predicts action-conditioned images or features}, world models~\cite{daydreamer,dreamerv3,gr1,gr2} roll out latent futures, and video-prediction policies~\cite{unipi,susie,avdc} generate or condition on pixel-level futures. EvoScene-VLA jointly denoises an action chunk and future scene states within the VLA action expert. The final scene state is carried into the next VLM prefix, where it is corrected using the new observation. This loop runs without a separate scene predictor at inference.

\paragraph{Flow matching and diffusion policies.}
Flow-matching and diffusion policies model robot control as trajectory denoising. Diffusion Policy~\cite{diffusion_policy} popularized diffusion-based continuous control. $\pi_0$~\cite{pi0}, RDT~\cite{rdt}, Octo~\cite{octo}, and flow-matching VLAs~\cite{flowmatching,rectified_flow} scale this formulation to pretrained VLAs. Consistency Policy~\cite{consistency_policy} accelerates sampling. These methods denoise actions only, so their decoders do not evolve scene state. EvoScene-VLA co-denoises the action chunk and the evolved scene representation in one flow-matching pass. This joint denoising grounds each action in the recurrent scene prior without adding an inference-time predictor.

\vspace{-1em}
\section{Method}
\vspace{-0.5em}

\subsection{Overview}
\vspace{-0.5em}
\begin{figure*}[t]
\centering
\includegraphics[width=1.02\textwidth]{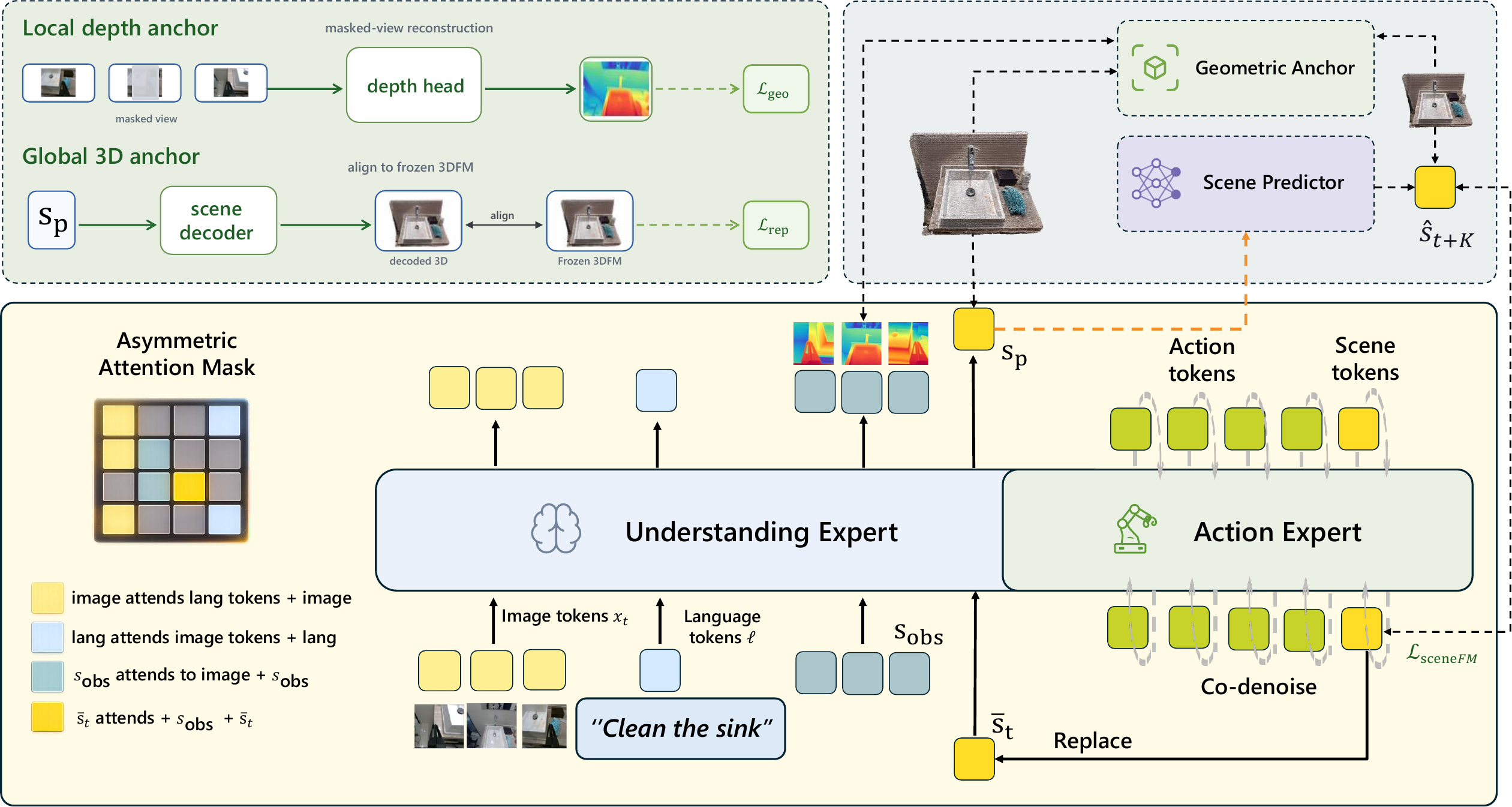}
\caption{\textbf{Pipeline Overview.} The policy receives multi-view images, an instruction, and the robot state. The VLM prefix contains image and language tokens, per-view observation slots, and recurrent prior slots.
Under asymmetric attention, observation slots attend to their corresponding image tokens, and prior slots attend to observation slots. Image and language tokens do not attend to scene slots.
During training, Geometric Anchor applies local depth supervision to per-view scene features and global 3DFM feature supervision to the fused state; Scene Predictor supplies future-state targets for the action expert.
At inference, the action expert jointly denoises the action chunk and future scene states at sparse offsets through flow matching.
The final state becomes the next call's prior, which the VLM fuses with the new observation to obtain the corrected current scene state.}
\label{fig:overview}
\vspace{-1em}
\end{figure*}

We build EvoScene-VLA on LingBot-VLA~\cite{lingbotvla}, a $\pi_{0.5}$-style~\cite{pi05} policy with a Qwen2.5-VL~\cite{qwen25vl} backbone.
In the base policy, the VLM encodes multi-view images and a language instruction into a key--value cache, which a separate action expert uses to generate an action chunk through flow matching.

EvoScene-VLA adds compact scene tokens to this interface (\cref{fig:overview}).
The action expert jointly denoises the action chunk and future scene states at sparse time steps, allowing predicted scene changes to inform action generation within the current chunk.
At inference, the state predicted for the end of the chunk serves as the prior for the next control call.
The VLM fuses this prior with the new observation to supplement incomplete visual evidence and correct the inherited state (\cref{sec:scenememory}).

To preserve control-relevant geometry, Geometric Anchor applies local depth supervision to per-view scene features and global 3D feature supervision to the fused scene state (\cref{sec:geo_anchor}).
Since demonstrations do not directly provide compact future scene states, Scene Predictor supplies targets for the action expert's scene flow-matching objective.
The predictor is supervised by 3DFM features extracted from future images (\cref{sec:scenepredictor}).
A training step is summarized in \cref{alg:train}.
At inference, we remove Geometric Anchor and Scene Predictor. Implementation details are provided in \cref{app:implementation}.
\vspace{-0.5em}

\subsection{Recurrent Scene-State Loop}
\label{sec:scenememory}
\label{sec:scenehead}
\vspace{-0.5em}

\paragraph{Scene representation and observation correction.}
Let $D$ denote the VLM hidden dimension and $V$ the number of camera views.
The scene interface contains $N_{\mathrm{obs}}$ observation slots per view, $s_{\mathrm{obs}}^{(v)} \in \mathbb{R}^{N_{\mathrm{obs}} \times D}$, and $N_{\mathrm{prior}}$ prior slots, $\bar{s}_t \in \mathbb{R}^{N_{\mathrm{prior}} \times D}$.
Together with the current multi-view images $x_t$ and language instruction $\ell$, they form the prefix $[\,x_t,\; \ell,\; s_{\mathrm{obs}}^{(1{:}V)},\; \bar{s}_t\,]$.
For the first chunk, $\bar{s}_t$ is initialized from learnable scene embeddings; at subsequent inference calls, it contains the final scene state generated for the preceding chunk.

The asymmetric attention mask in \cref{fig:overview} restricts each observation group to its own view's image tokens and observation slots, and the prior group to all observation and prior slots.
Image and language tokens do not attend to scene slots, preserving the pretrained vision--language attention paths.
The VLM therefore fuses current visual evidence with the prior through the per-view observation slots.
Its output at the prior-slot positions is the corrected current scene state $s_p \in \mathbb{R}^{N_{\mathrm{prior}} \times D}$:

\begin{equation}
    s_p
    =
    \mathrm{VLM}_{\mathrm{scene}}\bigl(x_t, \ell,\, s_{\mathrm{obs}}^{(1{:}V)},\, \bar{s}_t\bigr)_{\mathrm{prior}}
    \label{eq:scene_representation}
\end{equation}

\paragraph{Joint action and scene generation.}
Let $H$ denote the final action offset, so a chunk contains $H+1$ actions.
The action expert jointly generates this chunk and $K$ future scene states at sparse offsets $\{k_1,\ldots,k_K\} \subseteq \{1,\ldots,H\}$, with $k_K=H$.
At flow-matching time $\tau \in [0,1]$, let $a^\tau_{t{:}t+H}$ and $z^\tau \in \mathbb{R}^{K \times N_{\mathrm{prior}} \times D}$ denote the noisy action and scene blocks.
The expert receives the suffix $[\,r_t \;\big|\; z^\tau \;\big|\; a^\tau_{t{:}t+H}\,]$, where $r_t$ is the robot state, and attends to the VLM prefix cache.
Within the action expert, scene and action tokens attend to each other bidirectionally, allowing their predictions to inform each other at every denoising step.
The expert predicts a velocity $v_\theta$ for every action and scene token.
At inference, both blocks are initialized with independent Gaussian noise and jointly denoised using the same Euler sampling procedure as LingBot-VLA.
The outputs are an action chunk $\hat{a}_{t{:}t+H}$ and future scene states $\hat{z}^{(k_i)}$, $i=1,\ldots,K$.

\paragraph{State propagation across chunks.}
The final predicted state $\hat{z}^{(k_K)}$ corresponds to the time of the next observation.
After executing the action chunk, we pass this state to the next control call as a prior and combine it with the new observation $x_{t+1}$:
\begin{equation}
    \bar{s}_{t+1} \;=\; \hat{z}^{(k_K)},
    \qquad
    s_p^{\,t+1} \;=\; \mathrm{VLM}_{\mathrm{scene}}\bigl(x_{t+1}, \ell,\, s_{\mathrm{obs}}^{(1{:}V)},\, \bar{s}_{t+1}\bigr)_{\mathrm{prior}}
    \label{eq:recurrence}
\end{equation}
The new observation provides visual evidence to correct discrepancies caused by prediction errors or execution deviations.
\Cref{alg:infer} gives the inference procedure for one chunk.
The following sections describe the geometric supervision and future-state targets used to train this loop.

\subsection{Geometric Anchor}
\label{sec:supervision}
\label{sec:geo_anchor}
\label{sec:globalanchor}

To preserve control-relevant geometry, we apply a two-level \textbf{Geometric Anchor} during training: local depth supervision for per-view observation features $s_{\mathrm{obs}}^{(v)}$ and global 3D feature supervision for the fused state $s_p$.

\paragraph{Local depth supervision.}
For each target view $v_i$, the \textbf{Local depth anchor} masks its dense image tokens and predicts depth features from $s_{\mathrm{obs}}^{(v_i)}$, $s_p$, and the remaining views using a lightweight cross-attention head $g_{\mathrm{depth}}$:
\begin{equation}
    \hat{f}^{\,d}_{t,i}
    =
    g_{\mathrm{depth}}\!\left(
      q_{\mathrm{tmpl}},\;
      \bigl[\,
        \tilde{h}^{\mathrm{img},(i)}_{t,v_1},\ldots,\,
        \tilde{h}^{\mathrm{img},(i)}_{t,v_V},\,
        s_{\mathrm{obs}}^{(v_i)},\,
        s_p
      \,\bigr]
    \right),
    \quad
    \tilde{h}^{\mathrm{img},(i)}_{t,v_j}
    =
    \begin{cases}
      m, & j = i, \\
      h^{\mathrm{img}}_{t,v_j}, & j \neq i.
    \end{cases}
    \label{eq:depth_decoder}
\end{equation}
Here $q_{\mathrm{tmpl}}$ reuses LingBot-VLA's query tokens and $m$ is a learned mask embedding.
Target-view image information reaches the depth head through the scene representations.
A frozen Monocular Depth Teacher (MDT)~\cite{MORGBD} provides target features from the unmasked image:
\begin{equation}
    \mathcal{L}_{\mathrm{geo}}
    =
    \frac{1}{V}\sum_{i=1}^{V}
    \mathrm{SmoothL1}\!\left(
      \hat{f}^{\,d}_{t,i},\,
      f^{\,d}_{t,i}
    \right),
    \qquad
    f^{\,d}_{t,i} = \mathrm{MDT}(x_{t,v_i}).
    \label{eq:l_geo}
\end{equation}

\paragraph{Global 3D feature supervision.}
The \textbf{Global 3D anchor} supervises the fused state $s_p$ with features from a frozen multi-view 3DFM~\cite{pi3}.
A lightweight view-conditioned decoder $g_{3\mathrm{D}}$ takes learnable queries $q_{\mathrm{dec}}$ and uses $s_p$ as keys and values.
A linear projector $W_{\mathrm{proj}}$ maps the decoder output to the teacher feature space:
\begin{equation}
    H_{t} = g_{3\mathrm{D}}\!\left(q_{\mathrm{dec}};\, s_{p}\right),
    \qquad
    P_{t} = W_{\mathrm{proj}}\, H_{t}.
    \label{eq:global_decode}
\end{equation}
We align $P_t$ with 3DFM teacher features from the same multi-view images using an $\ell_1$ loss on per-token layer-normalized features:
\begin{equation}
    \mathcal{L}_{\mathrm{rep}}
    =
    \frac{1}{V}\sum_{v=1}^{V}
    \bigl\lVert \mathrm{LN}\bigl(P_{t}^{(v)}\bigr) - \mathrm{LN}\bigl(Z_{t}^{(v)}\bigr) \bigr\rVert_{1},
    \qquad
    Z_{t} = \mathrm{3DFM}(x_t).
    \label{eq:l_rep}
\end{equation}
Here $\mathrm{LN}$ is parameter-free layer normalization along the feature dimension.
The loss compares normalized features without directly matching their raw norms.

\paragraph{Shared geometric decoder.}
We reuse the global decoder $(g_{3\mathrm{D}}, W_{\mathrm{proj}})$ for the future states produced by Scene Predictor (\cref{sec:scenepredictor}).
Current and predicted future states are decoded into the same 3DFM feature space and supervised with teacher features from their respective time steps.

\subsection{Scene Predictor}
\label{sec:scenepredictor}

Demonstrations do not directly provide compact future scene states for the action expert's scene flow-matching objective.
We therefore train a \textbf{Scene Predictor} to produce these targets from the current scene state and demonstration actions.

\paragraph{Action-conditioned state prediction.}
The predictor is a causal Transformer with input sequence $[\,r_t,\; s_p,\; a_t,\ldots,a_{t+H},\; q_1,\ldots,q_K\,]$.
Each query group $q_i$ is initialized from $s_p$ and predicts a scene state at offset $k_i$.
Its attention mask allows access to the robot state $r_t$, the current scene state $s_p$, the demonstration action prefix $a_{t:t+k_i}$, and earlier query groups.
The output $\hat{s}_{t+k_i} \in \mathbb{R}^{N_{\mathrm{prior}} \times D}$ is thus conditioned on actions up to its target step.

\paragraph{Supervision from future observations.}
We extract teacher features $Z_{t+k_i}=\mathrm{3DFM}(x_{t+k_i})$ from future multi-view images.
The shared decoder $(g_{3\mathrm{D}}, W_{\mathrm{proj}})$ in \cref{eq:global_decode} maps each predicted state $\hat{s}_{t+k_i}$ to features $\tilde{P}_{t+k_i}$.
We train Scene Predictor with an $\ell_1$ loss against the future teacher features:
\begin{equation}
    \mathcal{L}_{\mathrm{pred}}
    =
    \frac{1}{K \cdot V}\sum_{i=1}^{K}\sum_{v=1}^{V}
    \bigl\lVert \tilde{P}_{t+k_i}^{(v)} - Z_{t+k_i}^{(v)} \bigr\rVert_{1},
    \qquad
    \tilde{P}_{t+k_i} = W_{\mathrm{proj}}\bigl(g_{3\mathrm{D}}(q_{\mathrm{dec}};\, \hat{s}_{t+k_i})\bigr).
    \label{eq:l_pred}
\end{equation}
These predicted states provide compact targets for the action expert's scene generation (\cref{sec:training_inference}).
During training, Scene Predictor also supplies the recurrent scene prior for chunks after the first.

\subsection{Training Objective}
\label{sec:training_inference}

\paragraph{Joint action and scene flow matching.}
We stack the future states predicted by Scene Predictor and apply $\mathrm{LN}$.
This gives standardized scene targets $z_0=\mathrm{LN}(\hat{s}_{t+k_1{:}t+k_K}) \in \mathbb{R}^{K \times N_{\mathrm{prior}} \times D}$ for training the action expert.
We sample a shared flow-matching time $\tau \in [0,1]$ for the action and scene blocks and draw independent standard Gaussian noises $\epsilon_a$ and $\epsilon_s$.
The noisy inputs to the joint generation process in \cref{sec:scenehead} are formed by linear interpolation:
\begin{equation}
    a^{\tau}_{t{:}t+H}
    =
    \tau\, \epsilon_a + (1-\tau)\, a_{t{:}t+H},
    \qquad
    z^{\tau}
    =
    \tau\, \epsilon_s + (1-\tau)\, z_0.
    \label{eq:scene_noise}
\end{equation}
We stop gradients through the action targets $a_{t{:}t+H}$ and scene targets $z_0$ for flow-matching training and optimize the velocity losses, where $\mathcal{L}_{\mathrm{actFM}}$ is the standard $\pi_{0.5}$ action flow-matching loss:
\begin{equation}
    \mathcal{L}_{\mathrm{sceneFM}}
    =
    \bigl\| v_{\theta}^{(s)}(z^{\tau},\tau) - (\epsilon_s - z_0) \bigr\|_2^2,
    \qquad
    \mathcal{L}_{\mathrm{actFM}}
    =
    \bigl\| v_{\theta}^{(a)}(a^{\tau},\tau) - (\epsilon_a - a_{t{:}t+H}) \bigr\|_2^2.
    \label{eq:l_scene_act}
\end{equation}

\paragraph{Full objective.}
The full objective combines action generation, geometric supervision, future-state prediction, and scene generation:
\begin{equation}
    \mathcal{L}
    =
    \mathcal{L}_{\mathrm{actFM}}
    +
    \lambda_{1}\mathcal{L}_{\mathrm{geo}}
    +
    \lambda_{2}\mathcal{L}_{\mathrm{rep}}
    +
    \lambda_{3}\mathcal{L}_{\mathrm{pred}}
    +
    \lambda_{4}\mathcal{L}_{\mathrm{sceneFM}}.
    \label{eq:total_loss}
\end{equation}
$\mathcal{L}_{\mathrm{geo}}$ and $\mathcal{L}_{\mathrm{rep}}$ supervise the geometry of current scene representations.
$\mathcal{L}_{\mathrm{pred}}$ supervises the future states produced by Scene Predictor, and $\mathcal{L}_{\mathrm{sceneFM}}$ trains the action expert using their standardized targets.
We jointly optimize the model with this objective in a single training stage.

\vspace{-2em}
\section{Experiments}
\vspace{-0.5em}
\subsection{Setup}

\paragraph{Benchmarks and baselines.}
We evaluate EvoScene-VLA on RoboTwin~\cite{robotwin}, LIBERO~\cite{liu2023libero}, and the Galaxea R1-Lite real robot (\cref{sec:real_robot}). On RoboTwin, we report main results on 31 tasks and conduct ablations on RoboTwin-5Task, which covers single-arm, dual-arm, short-horizon, and long-horizon tasks. We evaluate both RoboTwin sets under two settings. \emph{Clean} fixes initial object positions to the training distribution, whereas \emph{Rand} randomizes their positions and orientations within the task workspace while preserving the task goal. On LIBERO, we conduct component ablations across the Spatial, Object, Goal, and Long suites. We initialize EvoScene-VLA from the public LingBot-VLA~\cite{lingbotvla} checkpoint and compare it with the released LingBot-VLA policy and LingBot-VLA$^{*}$, which retains depth supervision during fine-tuning. All three models share the same backbone, chunk length, sampling schedule, and training protocol. We provide the exact task list in \cref{app:datasets}.

\vspace{-0.5em}
\subsection{Main Results}
\label{sec:main_results}

\paragraph{Quantitative results.}
\cref{tab:main_results} summarizes success rates on 31 RoboTwin tasks, with three-seed means for methods trained under our protocol. EvoScene-VLA raises the matched LingBot-VLA$^{*}$ average from 86.4\% to 88.8\% under \emph{Clean} and from 85.7\% to 87.9\% under \emph{Rand}. The paired gains ($\Delta$) are $2.4\pm0.4$ and $2.2\pm0.2$ percentage points, respectively (mean $\pm$ sample standard deviation across three seeds; \cref{tab:robotwin_full_results,app:multiseed}). EvoScene-VLA attains the higher mean on 22 of 31 tasks under \emph{Clean} and 23 of 31 under \emph{Rand}, so the gain is distributed across tasks and persists under randomized layouts. The published reference models ($\dagger$)~\cite{xvla,pi05,motus,rotvla} use 50-task training. Our averages exceed the reported Motus averages (87.0/85.8) and are numerically close to RotVLA (88.6/87.3) after re-averaging the published results over our 31 tasks.

\begin{figure}[t]
\centering
\captionsetup{font=small,skip=3pt}
\setlength{\tabcolsep}{0.75pt}
\newlength{\sceneprobeimagewidth}
\setlength{\sceneprobeimagewidth}{\dimexpr(\linewidth-12.5pt)/6\relax}
\newsavebox{\sceneprobeimagebox}
\sbox{\sceneprobeimagebox}{\includegraphics[width=\sceneprobeimagewidth]{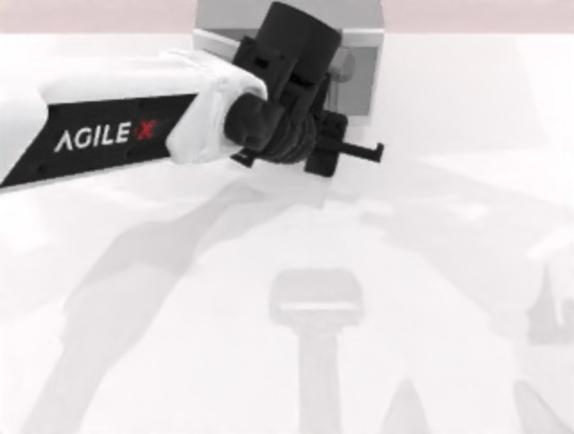}}
\begin{tabular}{@{}ccc@{\hspace{3pt}}c@{\hspace{3pt}}ccc@{}}
\multicolumn{3}{c}{\small\textit{Open Microwave}} & &
\multicolumn{3}{c}{\small\textit{Stack Blocks}} \\[0.2ex]
\scriptsize RGB & \scriptsize 3DFM reference & \scriptsize Decoded from $s_p$ & &
\scriptsize RGB & \scriptsize 3DFM reference & \scriptsize Decoded from $s_p$ \\[0.5ex]
\usebox{\sceneprobeimagebox} &
\includegraphics[width=\sceneprobeimagewidth]{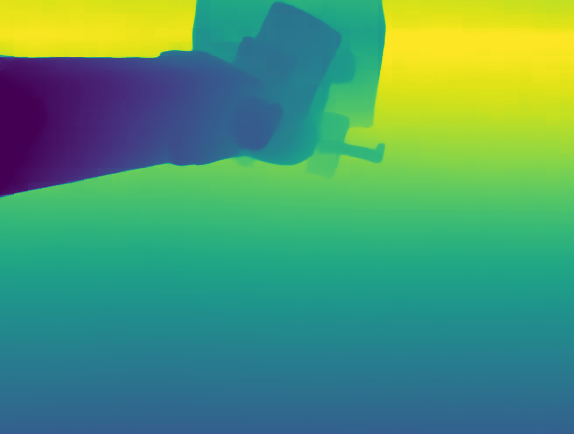} &
\includegraphics[width=\sceneprobeimagewidth]{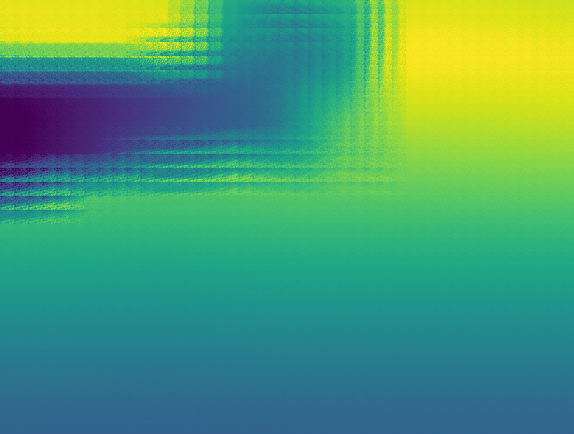} &
\textcolor{gray}{\vbox to \ht\sceneprobeimagebox{\offinterlineskip
\hrule width0.5pt height2.5pt
\vfill\hrule width0.5pt height2.5pt
\vfill\hrule width0.5pt height2.5pt
\vfill\hrule width0.5pt height2.5pt
\vfill\hrule width0.5pt height2.5pt
\vfill\hrule width0.5pt height2.5pt
\vfill\hrule width0.5pt height2.5pt
\vfill\hrule width0.5pt height2.5pt
\vfill\hrule width0.5pt height2.5pt
\vfill\hrule width0.5pt height2.5pt
}} &
\includegraphics[width=\sceneprobeimagewidth]{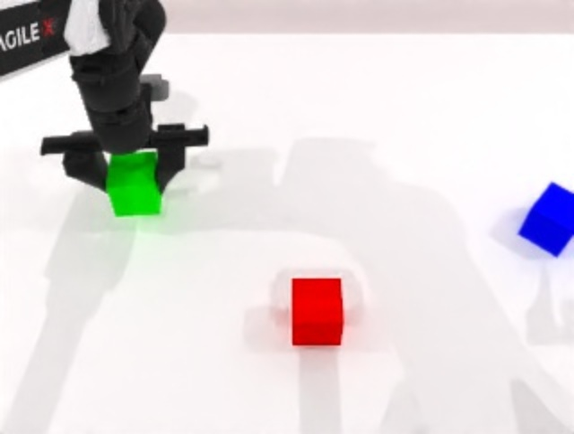} &
\includegraphics[width=\sceneprobeimagewidth]{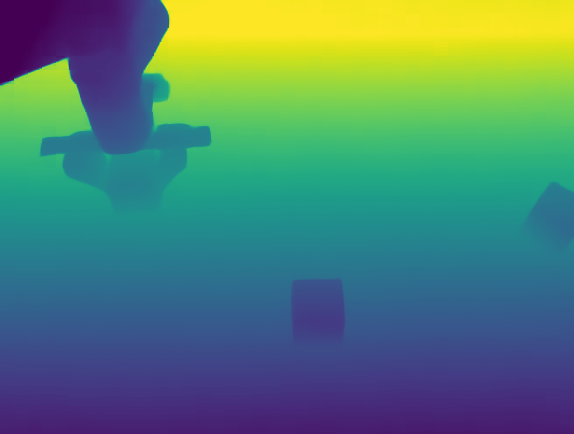} &
\includegraphics[width=\sceneprobeimagewidth]{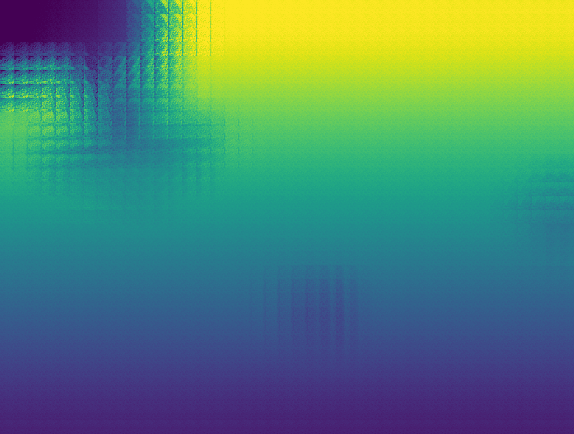}
\end{tabular}
\vspace{-1mm}
\caption{\textbf{Geometry decoded from the recurrent scene representation.} \emph{3DFM reference} decodes the frozen targets $Z_t$; \emph{Decoded from $s_p$} applies the same frozen point head to $P_t$ produced from $s_p$. Each triplet shares a reference-fitted colour range. The decoded maps retain coarse relative geometry while omitting fine details.}
\label{fig:scene_token_vis}
\vspace{-2mm}
\end{figure}

\begin{figure}[t]
\centering
\captionsetup{font=small,skip=3pt}
\setlength{\tabcolsep}{1pt}
\begin{tabular}{@{}cc@{}}
    \includegraphics[width=0.497\textwidth]{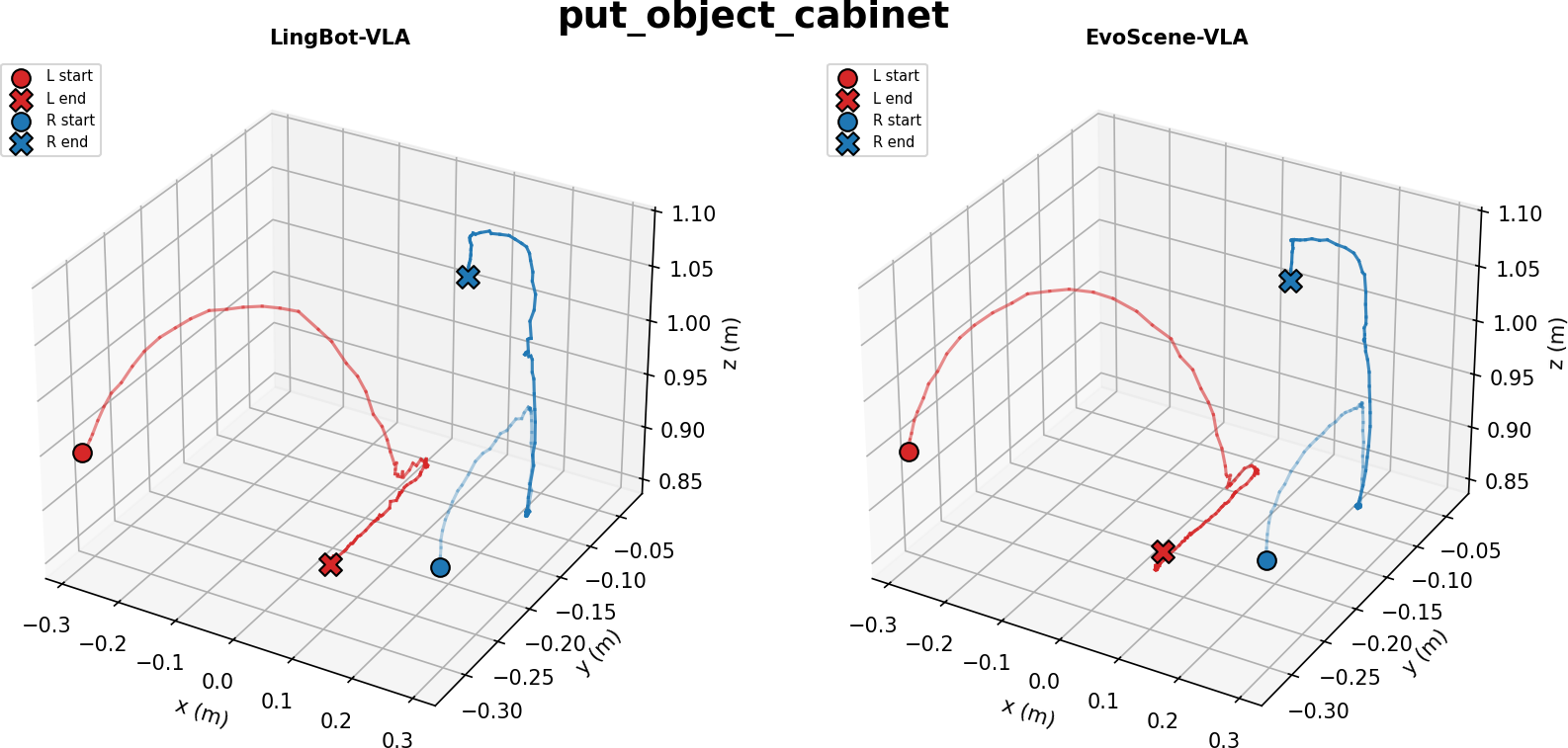} &
    \includegraphics[width=0.497\textwidth]{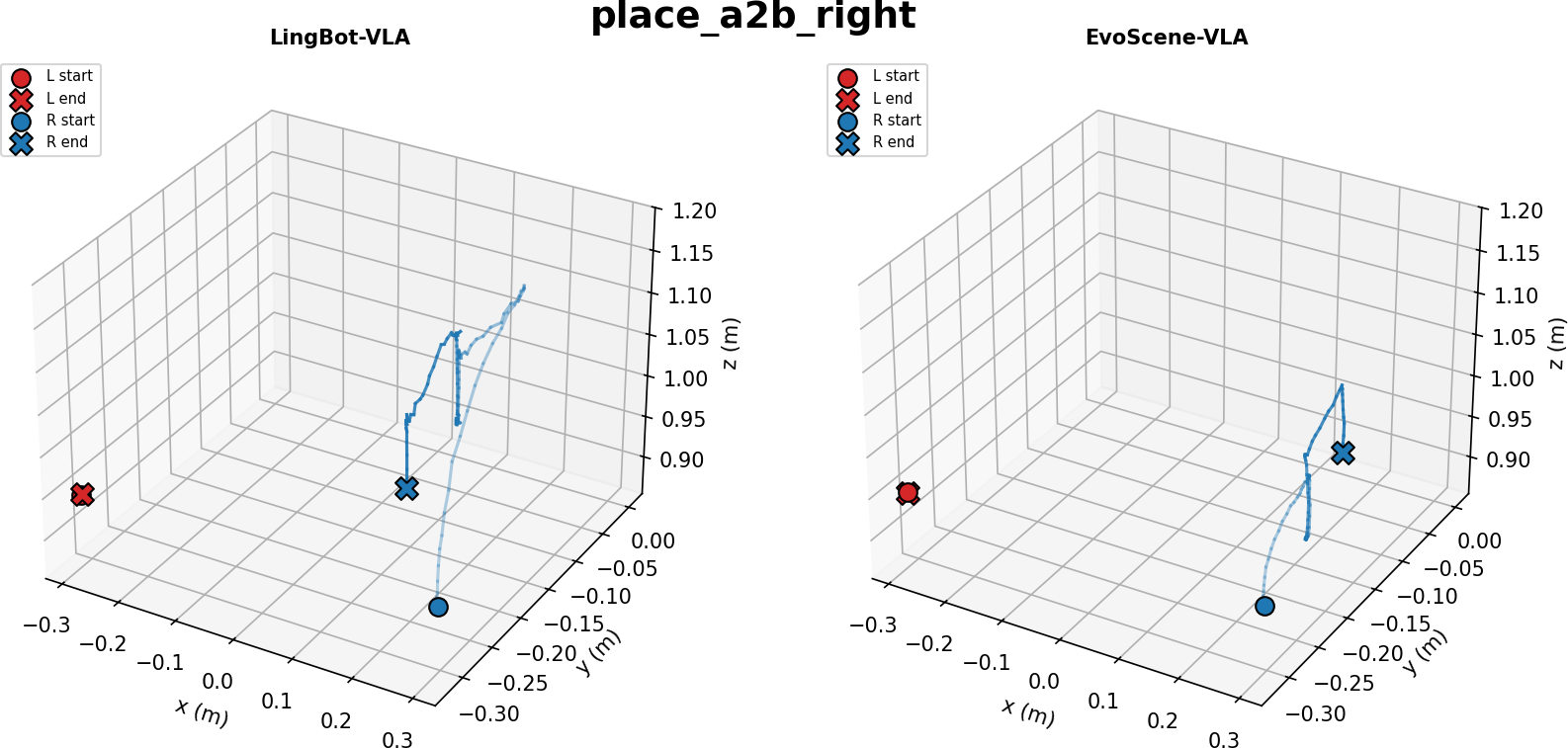}
\end{tabular}
\vspace{-1mm}

\caption{\textbf{End-effector trajectories in 3D.} EvoScene-VLA shows smoother paths than LingBot-VLA in these two representative RoboTwin episodes. \textcolor[HTML]{D62728}{Left}/\textcolor[HTML]{1F77B4}{right}-arm paths run from start (\mkstart) to end (\mkend).}

\label{fig:trajectory_plots}
\vspace{-1em}
\end{figure}

\vspace{-1em}
\paragraph{Qualitative results.}
\cref{fig:qualitative_analysis_a} examines the failure mode at the rollout level. For each of three tasks, the figure shows the LingBot-VLA$^{*}$ rollout next to ours from the same initial scene at three time slices (start, middle, end). In \emph{Open Microwave}, the baseline gripper retracts before reaching the door handle, whereas EvoScene-VLA continues the motion through to completion.
We also inspect the recurrent scene representation directly. \cref{fig:scene_token_vis} passes $P_t$, which the trained view-conditioned decoder produces from $s_p$, through the frozen 3DFM point head. The resulting maps preserve coarse relative geometry without fitting an additional probe.

\cref{fig:trajectory_plots} compares end-effector trajectories for representative RoboTwin episodes. EvoScene-VLA follows smoother paths in these examples. \Cref{app:trajectories} provides additional comparisons.

\begin{table}[t]
\hypersetup{hidelinks}
\centering
\captionsetup{font=small,skip=3pt}
\begin{minipage}[t]{0.47\textwidth}
\vspace{0pt}
\centering
\caption{\textbf{RoboTwin results.} Success rates (\%) on 31 tasks; Clean/Rand: fixed/randomized initial poses.}
\label{tab:main_results}
\footnotesize
\begingroup
\renewcommand{\arraystretch}{1.566}
\setlength{\tabcolsep}{2pt}
\begin{tabular*}{\linewidth}{@{\extracolsep{\fill}}lcc@{}}
\toprule
Method & Clean & Rand \\
\midrule
X-VLA$^{\dagger}$ & 70.7 & 71.1 \\
$\pi_{0.5}^{\dagger}$ & 81.2 & 75.9 \\
Motus$^{\dagger}$ & 87.0 & 85.8 \\
RotVLA$^{\dagger}$ & 88.6 & 87.3 \\
\midrule
LingBot-VLA & 84.6 & 83.6 \\
LingBot-VLA$^{*}$ & 86.4 & 85.7 \\
EvoScene-VLA & \textbf{88.8} & \textbf{87.9} \\
\midrule
$\Delta$ over LingBot-VLA$^{*}$ & $+2.4_{\pm0.4}$ & $+2.2_{\pm0.2}$ \\
\bottomrule
\end{tabular*}
\endgroup
\end{minipage}\hfill
\begin{minipage}[t]{0.50\textwidth}
\vspace{0pt}
\centering
\caption{\textbf{Relative-depth reconstruction.} LIBERO training frames with frozen 3DFM reference depth. Cold prior resets at each frame; the constant baseline predicts each image's median reference depth.}
\label{tab:scene_depth}
\footnotesize
\setlength{\tabcolsep}{2pt}
\begin{tabular*}{\linewidth}{@{\extracolsep{\fill}}lcccc@{}}
\toprule
Readout source & AbsRel $\downarrow$ & $\delta_1 \uparrow$ & $\delta_2 \uparrow$ & Pearson $\uparrow$ \\
\midrule
Recurrent prior & $\mathbf{0.163}$ & $\mathbf{0.732}$ & $\mathbf{0.930}$ & $\mathbf{0.831}$ \\
Cold prior & $0.172$ & $0.701$ & $0.919$ & $0.828$ \\
Constant baseline & $0.411$ & $0.586$ & $0.807$ & N/A \\
\bottomrule
\end{tabular*}

\par\vspace{4.5pt}
\caption{\textbf{Effect of prior content.} LIBERO success rates (\%). Matched priors: current episode; mismatched priors and replaced views: other episodes.}
\label{tab:prior_content}
\begingroup
\begin{tabular*}{\linewidth}{@{\extracolsep{\fill}}lccc@{}}
\toprule
Input condition & \makecell{Matched} & \makecell{Mismatched} & $\Delta$ (pp) \\
\midrule
unmodified-views & $85.45\%$ & $83.75\%$ & $+1.7$ \\
replaced Agent-view& $14.45\%$ & $1.65\%$ & $+12.8$ \\
replaced Wrist-view & $12.90\%$ & $2.70\%$ & $+10.2$ \\
\bottomrule
\end{tabular*}
\endgroup

\end{minipage}
\end{table}

\vspace{-1em}
\subsection{Ablations}
\vspace{-0.5em}
\paragraph{What the recurrent prior carries.}
Relative-depth reconstruction from the corrected current scene state $s_p$ is more accurate with a propagated prior than with a cold prior (\cref{tab:scene_depth}), with $\delta_1$ increasing from 0.701 to 0.732 against the frozen 3DFM reference.
We next compare closed-loop success with matched and mismatched priors (\cref{tab:prior_content,app:scene_depth}). Success is higher with matched priors in all three input conditions. The margin over mismatched priors ($\Delta$) is 1.7 percentage points (pp) with both original views, increasing to 12.8 and 10.2 points when the agent and wrist views are replaced, respectively. These results support the contribution of prior information.

\noindent\textbf{Cumulative component ablations.}\quad
\cref{tab:ablations} adds joint $\mathcal{L}_{\mathrm{pred}}$/$\mathcal{L}_{\mathrm{rep}}$ supervision, local depth anchoring, and recurrence on RoboTwin-5Task. Each step improves success over the preceding configuration, starting from a baseline without these components. LingBot-VLA$^{*}$ is included as a reference.

\noindent\textbf{Scene supervision and recurrence.}\quad
\cref{tab:libero_persuite} compares independent LIBERO configurations. The full model improves success by 2.70 points over the baseline. With full scene supervision, recurrence adds 1.10 points. Without joint $\mathcal{L}_{\mathrm{pred}}$/$\mathcal{L}_{\mathrm{rep}}$ supervision, recurrence reduces success by 2.10 points, or 2.35 points with local depth alone. Recurrence thus benefits from scene supervision.

\begingroup
\setlength{\intextsep}{2pt}
\vspace{3mm}
\begin{table}[H]
\centering
\captionsetup{font=small,skip=0pt}
\small
\setlength{\tabcolsep}{0pt}
\renewcommand{\arraystretch}{0.90}
\setlength{\extrarowheight}{0pt}
\begin{minipage}[t]{0.46\textwidth}
\centering
\caption{\textbf{Cumulative ablations.} Success rates (\%) averaged over the RoboTwin-5Task subset.}
\label{tab:ablations}
\setlength{\tabcolsep}{0pt}
\begin{tabular*}{\linewidth}{@{\extracolsep{\fill}}lcc@{}}
\toprule
Variant & Clean & Rand \\
\midrule
LingBot-VLA$^{*}$ & {87.8} & {84.6} \\
baseline & {81.6} & {75.8} \\
\midrule
+ $\mathcal{L}_{\mathrm{pred}}$ \& $\mathcal{L}_{\mathrm{rep}}$ & 89.3 & 86.2 \\
+ $\mathcal{L}_{\mathrm{geo}}$ & 90.1 & 86.5 \\
+ Recurrence & \textbf{90.8} & \textbf{87.8} \\
\bottomrule
\end{tabular*}
\end{minipage}
\hfill
\begin{minipage}[t]{0.50\textwidth}
\centering
\caption{\textbf{Real-robot experiments.} Success rates (\%) on three indoor-cleaning tasks with Galaxea R1-Lite.}
\label{tab:real_robot_results}
\setlength{\tabcolsep}{0pt}
\renewcommand{\arraystretch}{0.815}
\begin{tabular*}{\linewidth}{@{\extracolsep{\fill}}lcccc@{}}
\toprule
Task & $\pi_{0.5}$ & \makecell{LingBot\\VLA} & \makecell{LingBot\\VLA$^{*}$} & Ours \\
\midrule
Mirror & {28} & {27} & {26} & {29} \\
Sink & {42} & {44} & {49} & {51} \\
Cutting-board & {44} & {34} & {37} & {46} \\
\midrule
\textbf{Avg.$\uparrow$(\%)} & {38.0} & {35.0} & {37.3} & \textbf{42.0} \\
\bottomrule
\end{tabular*}
\end{minipage}
\end{table}

\vspace{3mm}
\newcommand{\liberodelta}[1]{%
  \begingroup
  \def\liberovalue{#1}%
  \def\liberodash{-}%
  \ifx\liberovalue\liberodash
    -%
  \else
    \ifdim#1pt>0pt
      \textcolor{red}{#1}%
    \else
      \ifdim#1pt<0pt
        \textcolor{blue}{#1}%
      \else
        \textcolor{black}{#1}%
      \fi
    \fi
  \fi
  \endgroup
}
\begin{table}[H]
\centering
\captionsetup{font=small,skip=0pt}
\caption{\textbf{Full factorial LIBERO component ablations.}
The $\checkmark$/$\times$ denote enabled/disabled components;
$\Delta$ is the overall change from the baseline. The 3DFM($\mathcal{L}_{\mathrm{pred}}$/$\mathcal{L}_{\mathrm{rep}}$)
directly supervises the recurrent state: without it, enabling recurrence
lowers success below the baseline ($\textcolor{blue}{-2.10}$), and $\mathcal{L}_{\mathrm{geo}}$ alone does not
recover it ($\textcolor{blue}{-1.50}$).}
\label{tab:libero_persuite}
\fontsize{8pt}{9.6pt}\selectfont
\setlength{\extrarowheight}{0pt}
\renewcommand{\arraystretch}{0.82}
\begin{tabular*}{\textwidth}{@{\extracolsep{\fill}}lccccccccc}
\toprule
\multirow{2}{*}{\textbf{Method}}
& \multicolumn{3}{c}{\textbf{Components}}
& \multicolumn{4}{c}{\textbf{LIBERO suite}}
& \multirow{2}{*}{\textbf{Overall}}
& \multirow{2}{*}{$\boldsymbol{\Delta}$} \\
\cmidrule(lr){2-4}\cmidrule(lr){5-8}
& $\mathcal{L}_{\mathrm{pred}}/\mathcal{L}_{\mathrm{rep}}$
& $\mathcal{L}_{\mathrm{geo}}$
& Recurrence
& \textbf{Spatial}
& \textbf{Object}
& \textbf{Goal}
& \textbf{Long} & & \\
\midrule
baseline
& $\times$ & $\times$ & $\times$
& 74.6 & 94.2 & 86.6 & 75.6 & 82.75 & \liberodelta{0.00} \\
\midrule
\multirow{7}{*}{EvoScene-VLA}
& $\checkmark$ & $\times$     & $\times$     & 74.4 & 92.6 & 90.2 & 76.2 & 83.35 & \liberodelta{+0.60} \\
& $\times$     & $\checkmark$ & $\times$     & \textbf{80.4} & 90.0 & 88.2 & 75.8 & 83.60 & \liberodelta{+0.85} \\
& $\times$     & $\times$     & $\checkmark$ & 72.4 & 89.2 & 86.2 & 74.8 & 80.65 & \liberodelta{-2.10} \\
& $\checkmark$ & $\checkmark$ & $\times$     & 79.4 & 93.4 & 90.2 & 74.4 & 84.35 & \liberodelta{+1.60} \\
& $\checkmark$ & $\times$     & $\checkmark$ & 79.2 & 91.8 & 90.4 & 76.8 & 84.55 & \liberodelta{+1.80} \\
& $\times$     & $\checkmark$ & $\checkmark$ & 72.6 & 90.2 & 87.2 & 75.0 & 81.25 & \liberodelta{-1.50} \\
& $\checkmark$ & $\checkmark$ & $\checkmark$ & 80.2 & \textbf{93.8} & \textbf{90.8} & \textbf{77.0} & \textbf{85.45} & \liberodelta{+2.70} \\
\bottomrule
\end{tabular*}
\end{table}

\endgroup

\FloatBarrier
\vspace{1em}
\subsection{Real-Robot Evaluation}
\label{sec:real_robot}

\begin{figure}[H]
\centering
\captionsetup{font=small,skip=3pt}

\begin{minipage}{\linewidth}
\begin{minipage}[t]{0.244\linewidth}
    \centering
    \includegraphics[width=\linewidth]{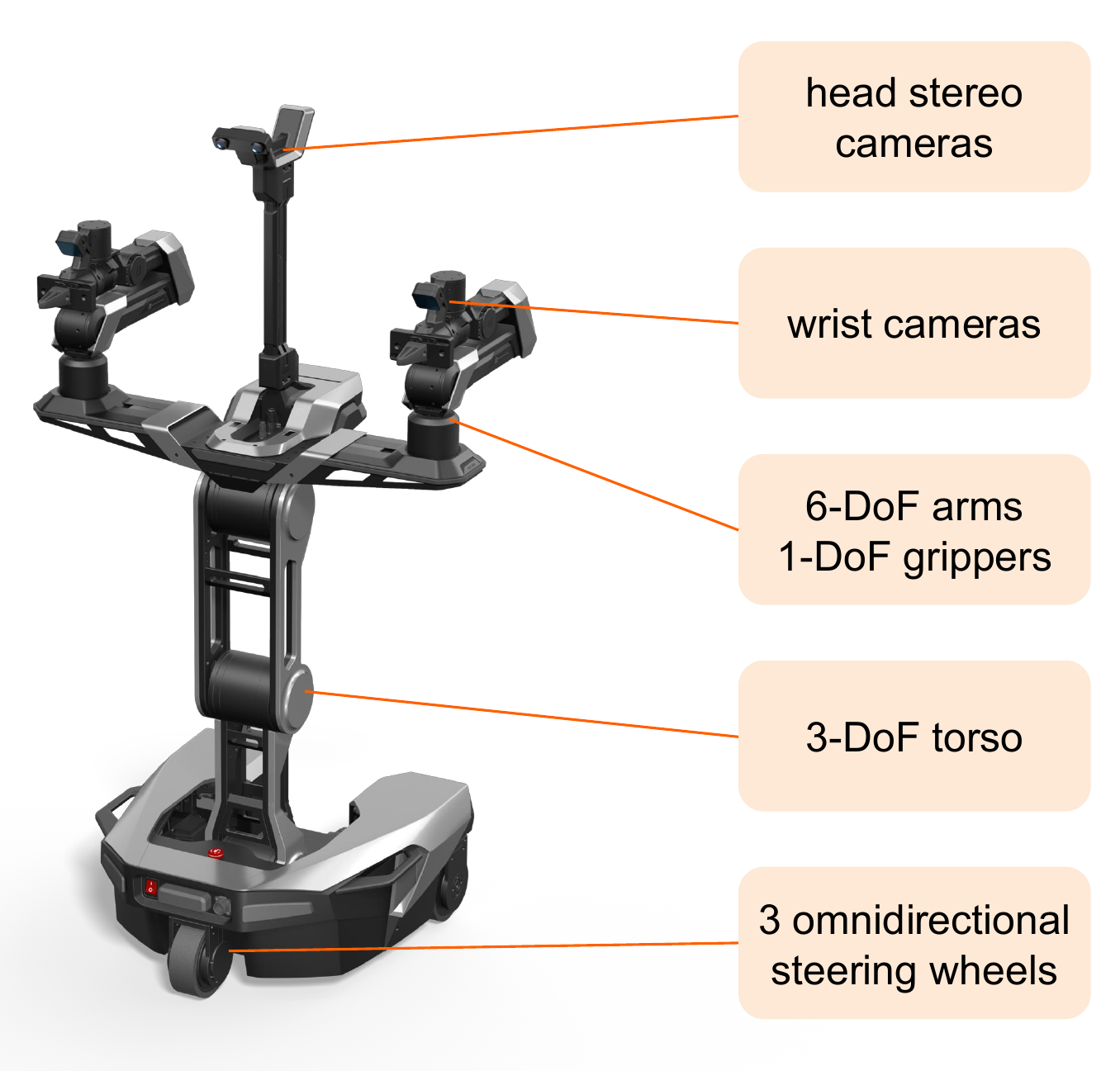}
    {\small (a) Robot platform.}
\end{minipage}
\hfill
\begin{minipage}[t]{0.75\linewidth}
    \centering
    \includegraphics[width=\linewidth]{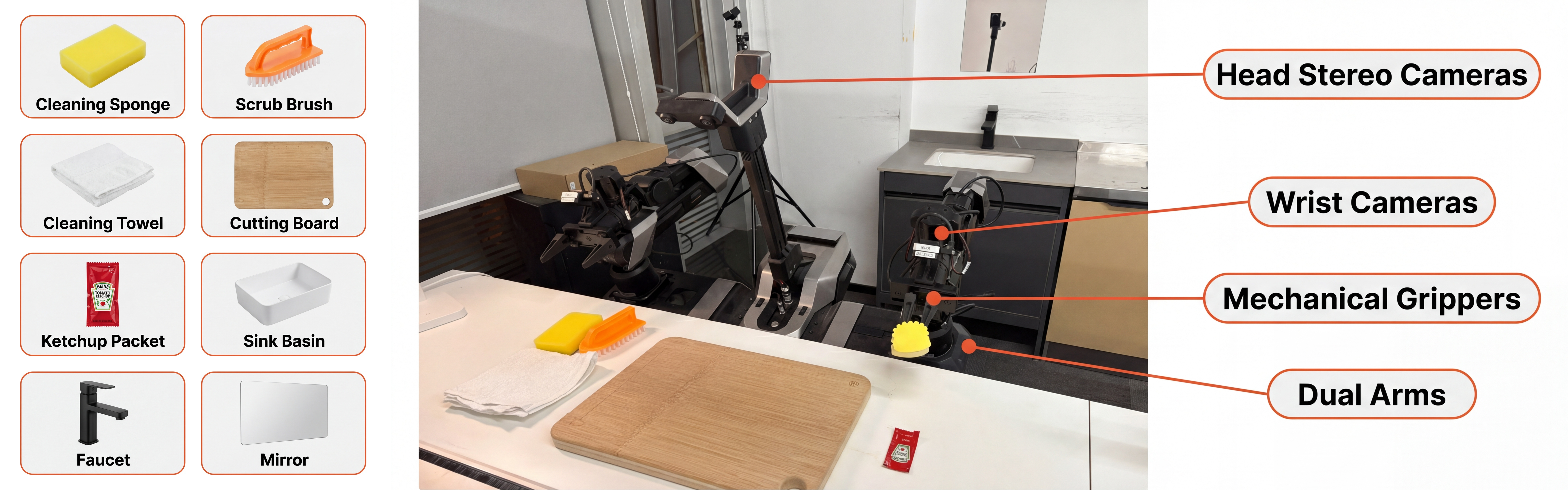}
    {\small (b) Real-robot experimental setup.}
\end{minipage}
\end{minipage}

\caption{
\textbf{Real-robot evaluation.} \textbf{(a)} Galaxea R1-Lite with three policy cameras.
\textbf{(b)} Indoor-cleaning setup for the mirror, sink, and cutting-board tasks; ketchup is the removable stain.
}
\label{fig:robot_setup}
\end{figure}

\paragraph{Platform and dataset.}
We evaluate EvoScene-VLA on the Galaxea R1-Lite platform (\cref{fig:robot_setup}(a)), training on the indoor-cleaning subset of the Galaxea Open-World Dataset~\cite{galaxea_openworld}, which includes mirror, sink, and cutting-board cleaning among other indoor tasks.
Following the standard VLA interface, the policy uses one head stream and two wrist streams as input. The subset contains 35 subdatasets, 2{,}669 episodes, and 3{,}667{,}408 frames, totaling approximately 67.9 hours of demonstrations at 15\,fps. We use all demonstrations for training and evaluate on the physical robot.

\vspace{-3mm}

\paragraph{Training and evaluation protocol.}
For the real-robot comparison, we train and deploy four methods: $\pi_{0.5}$, the two LingBot-VLA baselines, and EvoScene-VLA. We run all inference on a single NVIDIA RTX 4090 GPU. We evaluate each method over 100 physical closed-loop rollouts per task and report success rates under randomized object placements, lighting, and initial robot states.

\paragraph{Real-robot Results.}
The three long-horizon tasks shown in \cref{fig:robot_setup}(b) require the robot to track how its tool changes the surface. The surface state evolves during wiping, and subsequent controls need to target unwiped regions that can be subtle, occluded, or ambiguous in the current view.
\cref{tab:real_robot_results} reports success rates on the Galaxea cleaning tasks. EvoScene-VLA raises average success from 37.3\% for LingBot-VLA$^{*}$ to 42.0\% and achieves the highest observed success rate on each of the three tasks.
\cref{app:realrobot_ci} reports 95\% Wilson confidence intervals for every result. \cref{fig:realrobot_rollout_main} shows one closed-loop execution; \cref{app:realrobot_rollouts} presents its full sampled sequence alongside a failed rollout.

\begingroup
\setlength{\intextsep}{4pt}
\begin{figure}[H]
\centering
\captionsetup{font=small,skip=3pt}
\includegraphics[width=\linewidth]{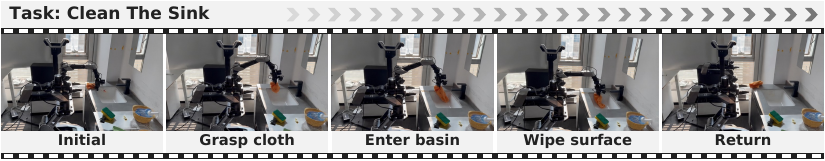}
\vspace{-1mm}
\caption{\textbf{Representative real-robot rollout on \emph{Clean The Sink}.} Frames progress from cloth acquisition to basin contact and wiping. The evolving interaction state motivates recurrent scene context.}
\label{fig:realrobot_rollout_main}
\vspace{-3mm}
\end{figure}
\endgroup

\section{Conclusion}

EvoScene-VLA uses compact scene tokens to unify within-chunk scene prediction, cross-chunk state propagation, and observation-based correction for chunked VLAs.
The action expert jointly denoises actions and future scene states in a single flow-matching process, allowing predicted scene changes to inform action generation.
The final predicted state becomes the next chunk's prior, which the VLM fuses with the new observation to correct the scene estimate.
During training, two-level Geometric Anchor provides geometric supervision, while Scene Predictor supplies future-state targets; both modules are removed at inference.
EvoScene-VLA improves average success over the matched baseline on 31 RoboTwin tasks, LIBERO, and three Galaxea R1-Lite real-robot cleaning tasks.
Ablations show that recurrence alone does not necessarily improve performance, whereas state propagation can improve control when combined with geometric and future scene supervision.

\bibliographystyle{iclr2027_conference}
\bibliography{references}

\subsection*{AI use statement}
We used generative AI tools to improve the English wording of this manuscript.
All suggested edits were reviewed by the authors, who take full responsibility for the final content.

\clearpage
\appendix
\raggedbottom
\captionsetup{font=small,skip=4pt}
\setlength{\textfloatsep}{10pt plus 2pt minus 2pt}
\setlength{\intextsep}{10pt plus 2pt minus 2pt}

\section*{Overview}

This appendix provides implementation and evaluation details. \cref{app:implementation} reports the optimization recipe, hyperparameters (\cref{tab:hparams}), and frozen-teacher pipeline. \cref{app:notation} (\cref{tab:notation}) consolidates the symbols used throughout the method, and \cref{app:algorithms} provides pseudocode for one training step (\cref{alg:train}) and one inference chunk (\cref{alg:infer}). \cref{app:datasets} lists the 31 RoboTwin tasks, the RoboTwin-5Task ablation subset, and baseline configurations. \revadd{\cref{app:multiseed} reports the per-task mean $\pm$ sample standard deviation across three seeds (\cref{tab:main_results_multiseed}), expanding \cref{tab:main_results} of the main paper, and \cref{app:realrobot_ci} reports 95\% Wilson binomial confidence intervals for the real-robot study (\cref{tab:real_robot_wilson_ci}). \locnew{\cref{app:realrobot_rollouts} shows two closed-loop real-robot rollouts (\cref{fig:realrobot_rollouts}).}} \Cref{app:trajectories} presents additional 3D end-effector trajectory comparisons (\cref{fig:appendix_trajectories}). \locnew{\revadd{\cref{app:robotwin_scene_decode} visualizes geometry decoded from the learned scene representation on RoboTwin (\cref{fig:robotwin_scene_decode}), and \cref{app:scene_depth} quantifies decoded depth and isolates the recurrent prior's contribution beyond the current observation (\cref{fig:scene_depth}, \cref{tab:scene_depth}, \cref{tab:prior_content}).}}

\section{Implementation}
\label[appendix]{app:implementation}

\paragraph{\revadd{Optimization} recipe.}
Following LingBot-VLA's recipe (\cref{tab:hparams}), the action expert is held in fp32 while the rest of the model runs in bf16 storage with fp32 reductions. The full model optimizes \cref{eq:total_loss}; the w/o Geometric Anchor variant drops the local and global anchor losses.

\paragraph{Hyperparameters.}
The loss weights $\lambda_1$--$\lambda_4$ in \cref{tab:hparams} are tuned on RoboTwin-5Task and reused on the full RoboTwin and Galaxea evaluations without further sweeping.

\begin{table}[h]
\centering
\caption{Hyperparameters introduced by EvoScene-VLA. \revadd{Optimization} values are inherited from the LingBot-VLA recipe; loss weights are EvoScene-VLA-specific.}
\label{tab:hparams}
\small
\renewcommand{\arraystretch}{0.94}
\begin{tabular}{ll}
\toprule
\textbf{\revadd{Optimization}} & \\
\midrule
\revadd{Optimizer} & AdamW \\
Learning rate (constant) & $1{\times}10^{-4}$ \\
Effective batch size & 256 \\
Total update steps & 20{,}000 \\
Mixed precision & bf16 storage / fp32 reductions \\
\midrule
\textbf{Architecture} & \\
\midrule
VLM backbone & Qwen2.5-VL-3B-Instruct \\
Hidden dimension $D$ & 2048 \\
Camera views $V$ & 3 (head, left wrist, right wrist) \\
Image resolution & $224{\times}224$ \\
Template bank size & 256 \\
Observation slots per view $N_{\mathrm{obs}}$ & 8 \\
Prior slots $N_{\mathrm{prior}}$ & 16 \\
Key frames $K$ & 3 \\
Action chunk length & 50 \\
\midrule
\textbf{Loss weights (\cref{eq:total_loss})} & \\
\midrule
$\lambda_1$ on $\mathcal{L}_{\mathrm{geo}}$ & 0.04 \\
$\lambda_2$ on $\mathcal{L}_{\mathrm{rep}}$ & 0.10 \\
$\lambda_3$ on $\mathcal{L}_{\mathrm{pred}}$ & 0.10 \\
$\lambda_4$ on $\mathcal{L}_{\mathrm{sceneFM}}$ & 0.01 \\
\midrule
\textbf{Inference} & \\
\midrule
Flow-matching Euler steps & 10 \\
Re-observation period & 1 chunk (50 steps) \\
\bottomrule
\end{tabular}
\end{table}

\paragraph{Frozen teachers.}
The local depth head $g_{\mathrm{depth}}$ reuses the LingBot-Depth alignment block, supervised by LingBot-VLA's MoRGBD pipeline~\cite{MORGBD} (MoGe-2 ViT-B normal variant combined with the LingBot-Depth masked-depth-modeling teacher) on the unmasked target view. The global 3D supervisor is the frozen Pi$^3$ multi-view 3D foundation model~\cite{pi3}, applied to current- and future-frame multi-view inputs for $\mathcal{L}_{\mathrm{rep}}$ and $\mathcal{L}_{\mathrm{pred}}$, respectively.

\clearpage
\section{Notation}
\label[appendix]{app:notation}

\cref{tab:notation} consolidates the symbols used throughout \cref{sec:scenememory,sec:geo_anchor,sec:scenepredictor,sec:scenehead}.

\begin{table}[h]
\centering
\caption{Notation used in EvoScene-VLA. Shapes are written as $\mathbb{R}^{\text{(dimensions)}}$ where applicable.}
\label{tab:notation}
\small
\begin{tabular}{ll}
\toprule
\textbf{Symbol} & \textbf{Meaning} \\
\midrule
\multicolumn{2}{l}{\textit{Indices and dimensions}} \\
$t$ & Discrete time index of the current control chunk \\
$D$ & VLM hidden dimension (Qwen2.5-VL-3B: $D{=}2048$) \\
$V$ & Number of camera views ($V{=}3$: head, left wrist, right wrist) \\
$N_{\mathrm{obs}}$ & Number of slots per observation group ($N_{\mathrm{obs}}{=}8$) \\
$N_{\mathrm{prior}}$ & Number of slots in the prior group ($N_{\mathrm{prior}}{=}16$) \\
$K$ & Number of scene key-frame query groups ($K{=}3$) \\
$H$ & Final action offset of the chunk; chunk length $H{+}1{=}50$ \\
$\{k_1,\ldots,k_K\}$ & Sparse key-frame offsets: $(1,24,49)$ in all experiments \\
\midrule
\multicolumn{2}{l}{\textit{Inputs}} \\
$x_t$ & Multi-view image input at step $t$ \\
$x_{t,v_i}$ & Image of view $i\in\{1,\ldots,V\}$ \\
$\ell$ & Language instruction \\
$r_t$ & Robot proprioceptive state \\
$a_{t{:}t+H}$ & Ground-truth action chunk \\
\midrule
\multicolumn{2}{l}{\textit{Recurrent scene prefix}} \\
$s_{\mathrm{obs}}^{(v)}\!\in\!\mathbb{R}^{N_{\mathrm{obs}}\times D}$ & Observation slots for view $v$, gather geometric evidence per camera \\
$\bar{s}_t\!\in\!\mathbb{R}^{N_{\mathrm{prior}}\times D}$ & Prior slots inherited from the previous chunk \\
$s_p\!\in\!\mathbb{R}^{N_{\mathrm{prior}}\times D}$ & VLM output at the prior-slot positions; cross-view scene representation \\
$h^{\mathrm{img}}_{t,v}$ & VLM image tokens for view $v$ \\
$m\in\mathbb{R}^D$ & Learned mask embedding for cross-view masking \\
\midrule
\multicolumn{2}{l}{\textit{Geometric Anchor (training only)}} \\
$q_{\mathrm{tmpl}}$ & 256-token template query bank reused for $g_{\mathrm{depth}}$ \\
$g_{\mathrm{depth}}$ & Cross-attention head producing per-view depth representations \\
$\hat{f}^{\,d}_{t,i}$ & Predicted depth representation for masked target view $i$ \\
$\mathrm{MDT}$ & Frozen Monocular Depth Teacher (MoGe-2 + LingBot-Depth) \\
$g_{3\mathrm{D}}$ & View-conditioned cross-attention decoder for the global anchor \\
$q_{\mathrm{dec}}$ & Learnable view-aware queries for $g_{3\mathrm{D}}$ \\
$W_{\mathrm{proj}}$ & Linear projector to the 3D foundation-model feature space \\
$P_t,\ Z_t$ & Projected output / frozen $\mathrm{3DFM}$ features for the current frame \\
\midrule
\multicolumn{2}{l}{\textit{Scene Predictor (training only)}} \\
$q_1,\ldots,q_K$ & Key-frame query groups initialized from $s_p$, each of size $N_{\mathrm{prior}}$ \\
$\hat{s}_{t+k_i}\!\in\!\mathbb{R}^{N_{\mathrm{prior}}\times D}$ & Predicted absolute future scene latent at offset $k_i$ \\
$\tilde{P}_{t+k_i},\ Z_{t+k_i}$ & Projected predicted feature / frozen $\mathrm{3DFM}$ feature on the future frame \\
\midrule
\multicolumn{2}{l}{\textit{Joint action--scene flow matching}} \\
$z_0\!\in\!\mathbb{R}^{K\times N_{\mathrm{prior}}\times D}$ & \revadd{Standardized} future-scene targets (LayerNorm of $\hat{s}$) \\
$\hat{z}^{(k_i)}\!\in\!\mathbb{R}^{N_{\mathrm{prior}}\times D}$ & Denoised standardized scene state emitted by the action expert at offset $k_i$ \\
$\tau\in[0,1]$ & Shared flow-matching time \\
$\epsilon_a,\ \epsilon_s$ & Independent Gaussian noises on action / scene paths \\
$a^{\tau}_{t{:}t+H},\ z^{\tau}$ & Straight-line interpolants \\
$v_{\theta}^{(a)},\ v_{\theta}^{(s)}$ & Predicted velocities for action / scene blocks \\
\midrule
\multicolumn{2}{l}{\textit{Losses (\cref{eq:total_loss})}} \\
$\mathcal{L}_{\mathrm{geo}}$ & Cross-view masked depth reconstruction loss \\
$\mathcal{L}_{\mathrm{rep}}$ & Global anchor $\ell_1$ loss for the current frame \\
$\mathcal{L}_{\mathrm{pred}}$ & Future-scene $\ell_1$ loss for the $K$ key frames \\
$\mathcal{L}_{\mathrm{sceneFM}},\ \mathcal{L}_{\mathrm{actFM}}$ & Scene / action flow-matching losses in the action expert \\
$\lambda_1,\lambda_2,\lambda_3,\lambda_4$ & Loss weights ($0.04,0.10,0.10,0.01$) \\
\bottomrule
\end{tabular}
\end{table}

\clearpage
\section{Algorithms}
\label[appendix]{app:algorithms}

\cref{alg:train} states one training step. \cref{alg:infer} states the deployed forward path.

\begin{algorithm}[h]
\caption{EvoScene-VLA training step.}
\label{alg:train}
\begin{algorithmic}[1]
\Require Sample $\bigl(x_t,\, \ell,\, r_t,\, a_{t{:}t+H},\, x_{t+k_{1}},\ldots,x_{t+k_{K}}\bigr)$ and prior carry $\bar{s}_t$ (learnable embedding for the first chunk in an episode; supplied by Scene Predictor for later chunks).
\State Form prefix $[\,x_t,\, \ell,\, s_{\mathrm{obs}}^{(1{:}V)},\, \bar{s}_t\,]$ with the asymmetric attention mask (\cref{sec:scenememory}).
\State Run VLM forward; read $s_p$ from the prior-slot positions (\cref{eq:scene_representation}).
\For{each view $i\in\{1,\ldots,V\}$} \Comment{Local depth anchor}
    \State Mask $h^{\mathrm{img}}_{t,v_i}\!\leftarrow\!m$, keep other views unchanged.
    \State $\hat{f}^{\,d}_{t,i}\leftarrow g_{\mathrm{depth}}(q_{\mathrm{tmpl}},\,[\,\tilde{h}^{\mathrm{img},(i)}_{t,v_1{:}V},\,s_{\mathrm{obs}}^{(v_i)},\,s_p\,])$
\EndFor
\State $\mathcal{L}_{\mathrm{geo}}\leftarrow \tfrac{1}{V}\sum_i \mathrm{SmoothL1}\!\bigl(\hat{f}^{\,d}_{t,i},\,\mathrm{MDT}(x_{t,v_i})\bigr)$ \Comment{\cref{eq:l_geo}}
\State $P_t\leftarrow W_{\mathrm{proj}}\,g_{3\mathrm{D}}(q_{\mathrm{dec}};\,s_p)$ \Comment{Global anchor}
\State $\mathcal{L}_{\mathrm{rep}}\leftarrow \tfrac{1}{V}\sum_v \bigl\lVert \mathrm{LN}\!\bigl(P_t^{(v)}\bigr)-\mathrm{LN}\!\bigl(\mathrm{3DFM}(x_t)^{(v)}\bigr)\bigr\rVert_{1}$
\State $\hat{s}_{t+k_{1}{:}t+k_{K}}\leftarrow \mathrm{ScenePredictor}(r_t,\,s_p,\,a_{t{:}t+H},\,q_1,\ldots,q_K)$
\State $\tilde{P}_{t+k_i}\leftarrow W_{\mathrm{proj}}\,g_{3\mathrm{D}}(q_{\mathrm{dec}};\,\hat{s}_{t+k_i})$ for each $i$
\State $\mathcal{L}_{\mathrm{pred}}\leftarrow \tfrac{1}{KV}\sum_{i,v}\bigl\lVert \tilde{P}_{t+k_i}^{(v)}-\mathrm{3DFM}(x_{t+k_i})^{(v)}\bigr\rVert_{1}$
\State $z_0\leftarrow \mathrm{LayerNorm}(\hat{s}_{t+k_{1}{:}t+k_{K}})$ \Comment{\revadd{Standardize} future-scene targets}
\State Sample $\tau\!\sim\!\mathcal{U}[0,1]$, $\epsilon_a,\epsilon_s\!\sim\!\mathcal{N}(0,I)$
\State $a^{\tau}\leftarrow \tau\epsilon_a + (1{-}\tau)a_{t{:}t+H}$,\quad $z^{\tau}\leftarrow \tau\epsilon_s + (1{-}\tau)z_0$
\State $v_{\theta}^{(a)},v_{\theta}^{(s)}\leftarrow \mathrm{ActionExpert}\bigl([\,r_t\,|\,z^{\tau}\,|\,a^{\tau}\,];\,\text{prefix cache}\bigr)$
\State $\mathcal{L}_{\mathrm{actFM}}\leftarrow \|v_{\theta}^{(a)}-(\epsilon_a-a_{t{:}t+H})\|_2^2$
\State $\mathcal{L}_{\mathrm{sceneFM}}\leftarrow \|v_{\theta}^{(s)}-(\epsilon_s-z_0)\|_2^2$
\State $\mathcal{L}\leftarrow \mathcal{L}_{\mathrm{actFM}}+\lambda_1\mathcal{L}_{\mathrm{geo}}+\lambda_2\mathcal{L}_{\mathrm{rep}}+\lambda_3\mathcal{L}_{\mathrm{pred}}+\lambda_4\mathcal{L}_{\mathrm{sceneFM}}$
\State Backpropagate $\mathcal{L}$; update VLM, action expert, and all training-only modules.
\end{algorithmic}
\end{algorithm}

\begin{algorithm}[h]
\caption{EvoScene-VLA inference per chunk.}
\label{alg:infer}
\begin{algorithmic}[1]
\Require Observation $x_t$, instruction $\ell$, robot state $r_t$, prior $\bar{s}_t$ (learnable first-chunk embedding).
\Ensure Action chunk $\hat{a}_{t{:}t+H}$ executed on the robot; updated prior $\bar{s}_{t+1}$.
\State Form prefix $[\,x_t,\, \ell,\, s_{\mathrm{obs}}^{(1{:}V)},\, \bar{s}_t\,]$ and run one VLM forward.
\State Read $s_p$ from prior slots; cache prefix KV. \Comment{Geometric Anchor / Scene Predictor not loaded.}
\State Initialise $a^{1}\!\sim\!\mathcal{N}(0,I)$, $z^{1}\!\sim\!\mathcal{N}(0,I)$.
\For{each Euler step $\tau\!\in\!\{1, 1{-}\Delta\tau,\ldots,\Delta\tau\}$ (10 steps total)}
    \State $v^{(a)},v^{(s)}\leftarrow \mathrm{ActionExpert}([\,r_t\,|\,z^{\tau}\,|\,a^{\tau}\,];\,\text{prefix cache})$
    \State $a^{\tau-\Delta\tau}\leftarrow a^{\tau}-\Delta\tau\, v^{(a)}$,\quad $z^{\tau-\Delta\tau}\leftarrow z^{\tau}-\Delta\tau\, v^{(s)}$
\EndFor
\State $\hat{a}_{t{:}t+H}\leftarrow a^{0}$;\quad $\bigl(\hat{z}^{(k_1)},\ldots,\hat{z}^{(k_K)}\bigr)\leftarrow z^0$ \Comment{Joint denoised outputs}
\State Execute $\hat{a}_{t{:}t+H}$ on the robot.
\State $\bar{s}_{t+1}\leftarrow \hat{z}^{(k_K)}$ \Comment{Write the denoised final-offset scene state back as the next prior}
\State Re-observe $x_{t+1}$ and return to step 1 for the next chunk.
\end{algorithmic}
\end{algorithm}

\FloatBarrier
\section{Datasets and Tasks}
\label[appendix]{app:datasets}

\paragraph{RoboTwin 31 tasks.}
We use 31 language-conditioned manipulation tasks from RoboTwin~2.0~\cite{robotwin}, covering single-arm and dual-arm, short-horizon and long-horizon settings. The full task list is: \texttt{place\_mouse\_pad}, \texttt{click\_bell}, \texttt{open\_microwave}, \texttt{place\_shoe}, \texttt{put\_object\_cabinet}, \texttt{stack\_blocks\_three}, \texttt{beat\_block\_hammer}, \texttt{turn\_switch}, \texttt{open\_laptop}, \texttt{place\_dual\_shoes}, \texttt{pick\_dual\_bottles}, \texttt{stack\_bowls\_three}, \texttt{place\_a2b\_left}, \texttt{place\_a2b\_right}, \texttt{place\_empty\_cup}, \texttt{move\_can\_pot}, \texttt{place\_container\_plate}, \texttt{press\_stapler}, \texttt{place\_phone\_stand}, \texttt{place\_fan}, \texttt{rotate\_qrcode}, \texttt{place\_object\_stand}, \texttt{shake\_bottle}, \texttt{scan\_object}, \texttt{pick\_diverse\_bottles}, \texttt{place\_bread\_skillet}, \texttt{place\_bread\_basket}, \texttt{place\_burger\_fries}, \texttt{place\_cans\_plasticbox}, \texttt{put\_bottles\_dustbin}, and \texttt{hanging\_mug}. Each task uses RoboTwin's standard episode length and the 50-step action chunk.

\paragraph{RoboTwin-5Task ablation subset.}
For ablations we use a 5-task subset that spans single-arm, dual-arm, short-horizon, and long-horizon types and keeps the ablation budget controlled: \texttt{click\_bell}, \texttt{open\_microwave}, \texttt{place\_shoe}, \texttt{put\_object\_cabinet}, and \texttt{stack\_blocks\_three}.

\paragraph{Baselines.}
LingBot-VLA$^{*}$ is the depth-augmented LingBot-VLA, corresponding to the \texttt{robotwin\_load20000h\_depth} configuration of the LingBot-VLA repository. EvoScene-VLA shares its training data, chunk length, and compute budget with the LingBot-VLA baselines. \revadd{The $\dagger$ rows of \cref{tab:main_results} are published results rather than our reproductions: the X-VLA and Motus values come from the per-task tables of Motus~\cite{motus} and the RotVLA values from RotVLA~\cite{rotvla}. Those models train jointly on all 50 RoboTwin~2.0 tasks with 50 clean and 500 randomized demonstrations per task; we report the unweighted mean over our 31 tasks.}

\Cref{tab:robotwin_full_results} gives the complete per-task comparison underlying the main-table averages.
\begin{table*}[!htbp]
\centering
\caption{Full per-task comparison corresponding to \cref{tab:main_results}. \revadd{Success rates (\%) on 31 RoboTwin tasks. C and R denote Clean and Rand. $\dagger$ marks published results re-averaged over the same tasks; other rows are three-seed means. Bold marks the column best. \textbf{Per-task mean $\pm$ sample standard deviation is in \cref{tab:main_results_multiseed}.}}}
\label{tab:robotwin_full_results}
\scriptsize
\setlength{\tabcolsep}{3.2pt}
\renewcommand{\arraystretch}{1.08}
\providecommand{\dwin}[1]{\textcolor{red}{#1}}
\providecommand{\dloss}[1]{\textcolor{blue}{#1}}

\resizebox{\textwidth}{!}{
\begin{tabular}{l|>{\columncolor{myblue}\centering\arraybackslash}p{2.4em}>{\columncolor{myblue}\centering\arraybackslash}p{2.4em} cc cc cc cc cc cc cc cc cc}
\toprule
\multirow{2}{*}{Method}
& \multicolumn{2}{|>{\columncolor{myblue}\centering\arraybackslash}p{\dimexpr4.8em+2\tabcolsep\relax}}{\makecell[c]{Avg.$\uparrow$\\(\%)}}
& \multicolumn{2}{c}{\makecell[c]{Place Mouse\\Pad}}
& \multicolumn{2}{c}{\makecell[c]{Click\\Bell}}
& \multicolumn{2}{c}{\makecell[c]{Open\\Microwave}}
& \multicolumn{2}{c}{\makecell[c]{Place\\Shoe}}
& \multicolumn{2}{c}{\makecell[c]{Put Obj.\\Cabinet}}
& \multicolumn{2}{c}{\makecell[c]{Stack Blocks\\Three}}
& \multicolumn{2}{c}{\makecell[c]{Beat Block\\Hammer}}
& \multicolumn{2}{c}{\makecell[c]{Turn\\Switch}}
& \multicolumn{2}{c}{\makecell[c]{Open\\Laptop}} \\
\cmidrule(lr){2-3}
\cmidrule(lr){4-5}
\cmidrule(lr){6-7}
\cmidrule(lr){8-9}
\cmidrule(lr){10-11}
\cmidrule(lr){12-13}
\cmidrule(lr){14-15}
\cmidrule(lr){16-17}
\cmidrule(lr){18-19}
\cmidrule(lr){20-21}
& C & R & C & R & C & R & C & R & C & R & C & R & C & R & C & R & C & R & C & R \\
\midrule
X-VLA$^{\dagger}$~\cite{xvla}
& 70.7 & 71.1
& 70 & 70
& \textbf{100} & \textbf{100}
& 79 & 71
& 96 & 95
& 46 & 48
& 6 & 10
& 92 & 88
& 40 & 61
& 93 & \textbf{100} \\

$\pi_{0.5}^{\dagger}$~\cite{pi05}
& 81.2 & 75.9
& 60 & 39
& 99 & 66
& 34 & 77
& 92 & 93
& 80 & 79
& 91 & 76
& \textbf{96} & \textbf{93}
& 62 & 54
& 90 & 96 \\

Motus$^{\dagger}$~\cite{motus}
& 87.0 & 85.8
& 66 & 68
& \textbf{100} & \textbf{100}
& \textbf{95} & \textbf{91}
& \textbf{99} & 97
& \textbf{88} & 71
& 91 & \textbf{95}
& 95 & 88
& \textbf{84} & \textbf{78}
& 95 & 91 \\

RotVLA$^{\dagger}$~\cite{rotvla}
& 88.6 & 87.3
& 83 & 87
& 97 & 92
& 36 & 12
& \textbf{99} & \textbf{100}
& 87 & 82
& 90 & 86
& 92 & 85
& 46 & 46
& 98 & 99 \\

\midrule

LingBot-VLA
& 84.6 & 83.6
& 84 & 88
& 68 & 60
& 66 & 54
& 97 & 95
& 82 & 80
& 95 & 90
& 90 & 85
& 64 & 75
& 98 & 97 \\

LingBot-VLA$^{*}$
& 86.4 & 85.7
& 84 & 87
& 95 & 96
& 70 & 58
& 97 & 98
& 82 & 83
& 95 & 88
& 87 & 83
& 71 & 71
& 97 & 98 \\

Ours
& \textbf{88.8} & \textbf{87.9}
& \textbf{85} & \textbf{91}
& 96 & 97
& 82 & 61
& 95 & 99
& 85 & \textbf{87}
& \textbf{96} & \textbf{95}
& 93 & 84
& 72 & 75
& \textbf{99} & 99 \\

$\Delta$
& \dwin{+2.4} & \dwin{+2.2}
& \dwin{+1} & \dwin{+4}
& \dwin{+1} & \dwin{+1}
& \dwin{+12} & \dwin{+3}
& \dloss{-2} & \dwin{+1}
& \dwin{+3} & \dwin{+4}
& \dwin{+1} & \dwin{+7}
& \dwin{+6} & \dwin{+1}
& \dwin{+1} & \dwin{+4}
& \dwin{+2} & \dwin{+1} \\
\bottomrule
\end{tabular}
}

\vspace{-0.1em}

\resizebox{\textwidth}{!}{
\begin{tabular}{cc cc cc cc cc cc cc cc cc cc cc}
\toprule
\multicolumn{2}{c}{\makecell[c]{Place Dual\\Shoes}}
& \multicolumn{2}{c}{\makecell[c]{Pick Dual\\Bottles}}
& \multicolumn{2}{c}{\makecell[c]{Stack Bowls\\Three}}
& \multicolumn{2}{c}{\makecell[c]{Place A2B\\Left}}
& \multicolumn{2}{c}{\makecell[c]{Place A2B\\Right}}
& \multicolumn{2}{c}{\makecell[c]{Place Empty\\Cup}}
& \multicolumn{2}{c}{\makecell[c]{Move Can\\Pot}}
& \multicolumn{2}{c}{\makecell[c]{Place Cont.\\Plate}}
& \multicolumn{2}{c}{\makecell[c]{Press\\Stapler}}
& \multicolumn{2}{c}{\makecell[c]{Place Phone\\Stand}}
& \multicolumn{2}{c}{\makecell[c]{Place\\Fan}} \\
\cmidrule(lr){1-2}
\cmidrule(lr){3-4}
\cmidrule(lr){5-6}
\cmidrule(lr){7-8}
\cmidrule(lr){9-10}
\cmidrule(lr){11-12}
\cmidrule(lr){13-14}
\cmidrule(lr){15-16}
\cmidrule(lr){17-18}
\cmidrule(lr){19-20}
\cmidrule(lr){21-22}
C & R & C & R & C & R & C & R & C & R & C & R & C & R & C & R & C & R & C & R & C & R \\
\midrule
79 & 88
& 47 & 36
& 76 & 86
& 48 & 49
& 36 & 36
& \textbf{100} & 98
& 89 & 86
& 97 & 95
& 92 & \textbf{98}
& 88 & 87
& 80 & 75 \\

75 & 75
& 93 & 63
& 77 & 71
& 87 & 82
& 87 & 84
& \textbf{100} & 99
& 51 & 55
& 99 & 95
& 87 & 83
& 81 & 81
& 87 & 85 \\

\textbf{93} & 87
& \textbf{96} & \textbf{90}
& 79 & 87
& 88 & 79
& 91 & 87
& 99 & 98
& 34 & 74
& 98 & 99
& \textbf{93} & \textbf{98}
& 87 & 86
& 91 & 87 \\

\textbf{93} & \textbf{96}
& 89 & 88
& \textbf{92} & \textbf{88}
& \textbf{93} & \textbf{95}
& \textbf{97} & \textbf{96}
& \textbf{100} & 99
& \textbf{95} & \textbf{95}
& 99 & \textbf{100}
& 83 & 81
& \textbf{94} & 94
& \textbf{97} & \textbf{97} \\

\midrule

90 & 89
& 88 & 85
& 72 & 76
& 76 & 80
& 77 & 70
& \textbf{100} & \textbf{100}
& 75 & 66
& 96 & 98
& 90 & 95
& 84 & \textbf{96}
& 94 & 89 \\

75 & 87
& 93 & 88
& 75 & 82
& 79 & 77
& 76 & 76
& \textbf{100} & 99
& 82 & 90
& 95 & 96
& 92 & 94
& 91 & 91
& 93 & 86 \\

91 & 93
& 87 & 88
& 83 & 76
& 80 & 83
& 81 & 79
& \textbf{100} & \textbf{100}
& 75 & 75
& \textbf{100} & 98
& 92 & 95
& 91 & \textbf{96}
& \textbf{97} & 91 \\

\dwin{+16} & \dwin{+6}
& \dloss{-6} & +0
& \dwin{+8} & \dloss{-6}
& \dwin{+1} & \dwin{+6}
& \dwin{+5} & \dwin{+3}
& +0 & \dwin{+1}
& \dloss{-7} & \dloss{-15}
& \dwin{+5} & \dwin{+2}
& +0 & \dwin{+1}
& +0 & \dwin{+5}
& \dwin{+4} & \dwin{+5} \\
\bottomrule
\end{tabular}
}

\vspace{-0.1em}

\resizebox{\textwidth}{!}{
\begin{tabular}{cc cc cc cc cc cc cc cc cc cc cc}
\toprule
\multicolumn{2}{c}{\makecell[c]{Rotate\\QRcode}}
& \multicolumn{2}{c}{\makecell[c]{Place Obj.\\Stand}}
& \multicolumn{2}{c}{\makecell[c]{Shake\\Bottle}}
& \multicolumn{2}{c}{\makecell[c]{Scan\\Obj.}}
& \multicolumn{2}{c}{\makecell[c]{Pick Diverse\\Bottles}}
& \multicolumn{2}{c}{\makecell[c]{Place Bread\\Skillet}}
& \multicolumn{2}{c}{\makecell[c]{Place Bread\\Basket}}
& \multicolumn{2}{c}{\makecell[c]{Place Burger\\Fries}}
& \multicolumn{2}{c}{\makecell[c]{Place Cans\\Plasticbox}}
& \multicolumn{2}{c}{\makecell[c]{Put Bottles\\Dustbin}}
& \multicolumn{2}{c}{\makecell[c]{Hanging\\Mug}} \\
\cmidrule(lr){1-2}
\cmidrule(lr){3-4}
\cmidrule(lr){5-6}
\cmidrule(lr){7-8}
\cmidrule(lr){9-10}
\cmidrule(lr){11-12}
\cmidrule(lr){13-14}
\cmidrule(lr){15-16}
\cmidrule(lr){17-18}
\cmidrule(lr){19-20}
\cmidrule(lr){21-22}
C & R & C & R & C & R & C & R & C & R & C & R & C & R & C & R & C & R & C & R & C & R \\
\midrule
34 & 33
& 86 & 88
& 99 & \textbf{100}
& 14 & 36
& 58 & 36
& 77 & 67
& 81 & 71
& 94 & 94
& 97 & 98
& 74 & 77
& 23 & 27 \\

89 & 87
& 91 & 85
& 99 & 97
& 18 & 17
& 81 & 71
& 85 & 66
& 77 & 64
& 94 & 87
& 94 & 84
& 84 & 79
& 77 & 71 \\

89 & 73
& \textbf{98} & 97
& \textbf{100} & 97
& 67 & 66
& \textbf{90} & \textbf{91}
& 86 & 83
& 91 & \textbf{94}
& 98 & 98
& 98 & 94
& 81 & 79
& 38 & 38 \\

\textbf{95} & \textbf{95}
& 97 & \textbf{100}
& \textbf{100} & \textbf{100}
& \textbf{88} & \textbf{85}
& 88 & 88
& 89 & 88
& \textbf{96} & 93
& \textbf{99} & 98
& \textbf{100} & \textbf{100}
& 91 & 88
& 44 & 52 \\

\midrule

78 & 86
& 96 & 90
& \textbf{100} & 99
& 57 & 47
& 78 & 69
& \textbf{91} & \textbf{91}
& 90 & 90
& 96 & 96
& 98 & 98
& 80 & 83
& 72 & 76 \\

83 & 87
& 96 & 85
& 98 & 99
& 56 & 38
& 80 & 74
& 89 & \textbf{91}
& 91 & 91
& 95 & \textbf{99}
& 99 & 96
& 88 & 87
& 75 & \textbf{82} \\

88 & 87
& \textbf{98} & 96
& \textbf{100} & \textbf{100}
& 49 & 57
& 83 & 80
& 89 & 87
& 92 & 91
& \textbf{99} & 97
& 99 & 99
& \textbf{94} & \textbf{93}
& \textbf{83} & 76 \\

\dwin{+5} & +0
& \dwin{+2} & \dwin{+11}
& \dwin{+2} & \dwin{+1}
& \dloss{-7} & \dwin{+19}
& \dwin{+3} & \dwin{+6}
& +0 & \dloss{-4}
& \dwin{+1} & +0
& \dwin{+4} & \dloss{-2}
& +0 & \dwin{+3}
& \dwin{+6} & \dwin{+6}
& \dwin{+8} & \dloss{-6} \\
\bottomrule
\end{tabular}
}

\end{table*}

\clearpage
\revadd{\section{Per-Task Multi-Seed Results}}
\label[appendix]{app:multiseed}

\revadd{\cref{tab:main_results_multiseed} expands \cref{tab:main_results} with the
per-task uncertainty. For the LingBot-VLA baselines and EvoScene-VLA we trained
three independent seeds and evaluated each with 100 rollouts per task and setting, and
we report the mean $\pm$ sample standard deviation across those three seeds. The
$\Delta$ row gives the paired per-task improvement of EvoScene-VLA over
LingBot-VLA$^{*}$ together with the standard deviation of that difference. Aggregating
over the 31 tasks, the macro averages are $86.4\pm0.7$ versus $88.8\pm0.2$ under Clean
($+2.4\pm0.4$) and $85.7\pm0.3$ versus $87.9\pm0.5$ under Rand ($+2.2\pm0.2$).
EvoScene-VLA attains the higher mean on 22 of 31 tasks under Clean and 23 of 31 under
Rand, so the gain is broadly distributed rather than concentrated in a few tasks.}

\begin{table}[H]
\centering
\caption{\revadd{\textbf{Per-task success rates on the 31 RoboTwin tasks under the Clean and Rand settings} (\%). Expanded version of \cref{tab:main_results}. Each cell is the mean $\pm$ sample standard deviation over three independently trained seeds, with 100 rollouts per task, setting, and seed. The $\Delta$ columns give the paired improvement of EvoScene-VLA over LingBot-VLA$^{*}$, with the standard deviation of that difference across the three seeds. Per-row best per setting is bolded among the three methods.}}
\label{tab:main_results_multiseed}
\scriptsize
\setlength{\tabcolsep}{1.5pt}
\renewcommand{\arraystretch}{1.4}
\providecommand{\seedms}[2]{\ensuremath{#1\,\pm\,#2}}
\providecommand{\bestseedms}[2]{\ensuremath{\mathbf{#1}\,\pm\,#2}}
\providecommand{\seeddeltams}[2]{%
  \begingroup
  \ifdim#1pt>0pt
    \color{red}%
  \else
    \ifdim#1pt<0pt
      \color{blue}%
    \else
      \color{black}%
    \fi
  \fi
  \seedms{#1}{#2}%
  \endgroup
}
\resizebox{\textwidth}{!}{%
\begin{tabular}{@{}l cc|cc|cc|cc@{}}
\toprule
& \multicolumn{2}{c|}{LingBot-VLA} & \multicolumn{2}{c|}{LingBot-VLA$^{*}$} & \multicolumn{2}{c|}{\textbf{EvoScene-VLA (Ours)}} & \multicolumn{2}{c}{$\Delta$} \\
\cmidrule(lr){2-3}\cmidrule(lr){4-5}\cmidrule(lr){6-7}\cmidrule(lr){8-9}
\textbf{Task} & Clean & Rand. & Clean & Rand. & Clean & Rand. & Clean & Rand. \\
\midrule
Place Mouse Pad & \seedms{84.0}{2.6} & \seedms{88.0}{4.6} & \seedms{84.0}{3.6} & \seedms{87.0}{6.0} & \bestseedms{85.0}{4.6} & \bestseedms{91.0}{4.0} & \seeddeltams{+1.0}{1.0} & \seeddeltams{+4.0}{5.3} \\
Click Bell & \seedms{68.0}{5.3} & \seedms{60.0}{6.0} & \seedms{95.0}{5.3} & \seedms{96.0}{2.6} & \bestseedms{96.0}{1.7} & \bestseedms{97.0}{2.6} & \seeddeltams{+1.0}{4.4} & \seeddeltams{+1.0}{5.2} \\
Open Microwave & \seedms{66.0}{5.6} & \seedms{54.0}{5.0} & \seedms{70.0}{6.0} & \seedms{58.0}{3.0} & \bestseedms{82.0}{2.6} & \bestseedms{61.0}{1.7} & \seeddeltams{+12.0}{8.2} & \seeddeltams{+3.0}{4.6} \\
Place Shoe & \bestseedms{97.0}{2.0} & \seedms{95.0}{5.0} & \bestseedms{97.0}{1.0} & \seedms{98.0}{1.7} & \seedms{95.0}{3.0} & \bestseedms{99.0}{1.0} & \seeddeltams{-2.0}{3.6} & \seeddeltams{+1.0}{2.6} \\
Put Object Cabinet & \seedms{82.0}{1.0} & \seedms{80.0}{3.6} & \seedms{82.0}{2.0} & \seedms{83.0}{3.0} & \bestseedms{85.0}{5.6} & \bestseedms{87.0}{4.6} & \seeddeltams{+3.0}{3.6} & \seeddeltams{+4.0}{6.9} \\
Stack Blocks Three & \seedms{95.0}{1.7} & \seedms{90.0}{4.0} & \seedms{95.0}{2.0} & \seedms{88.0}{4.4} & \bestseedms{96.0}{2.6} & \bestseedms{95.0}{4.6} & \seeddeltams{+1.0}{4.4} & \seeddeltams{+7.0}{3.6} \\
Beat Block Hammer & \seedms{90.0}{6.0} & \bestseedms{85.0}{3.6} & \seedms{87.0}{5.0} & \seedms{83.0}{4.6} & \bestseedms{93.0}{6.0} & \seedms{84.0}{6.0} & \seeddeltams{+6.0}{9.5} & \seeddeltams{+1.0}{9.6} \\
Turn Switch & \seedms{64.0}{2.6} & \bestseedms{75.0}{5.3} & \seedms{71.0}{3.5} & \seedms{71.0}{5.6} & \bestseedms{72.0}{3.6} & \bestseedms{75.0}{2.6} & \seeddeltams{+1.0}{2.6} & \seeddeltams{+4.0}{3.5} \\
Open Laptop & \seedms{98.0}{1.7} & \seedms{97.0}{5.2} & \seedms{97.0}{1.0} & \seedms{98.0}{1.0} & \bestseedms{99.0}{1.0} & \bestseedms{99.0}{1.0} & \seeddeltams{+2.0}{1.0} & \seeddeltams{+1.0}{1.0} \\
Place Dual Shoes & \seedms{90.0}{4.6} & \seedms{89.0}{2.6} & \seedms{75.0}{4.6} & \seedms{87.0}{5.0} & \bestseedms{91.0}{3.6} & \bestseedms{93.0}{5.3} & \seeddeltams{+16.0}{4.0} & \seeddeltams{+6.0}{7.9} \\
Pick Dual Bottles & \seedms{88.0}{2.6} & \seedms{85.0}{2.0} & \bestseedms{93.0}{2.6} & \bestseedms{88.0}{2.6} & \seedms{87.0}{5.3} & \bestseedms{88.0}{3.5} & \seeddeltams{-6.0}{7.8} & \seeddeltams{+0.0}{1.0} \\
Stack Bowls Three & \seedms{72.0}{4.0} & \seedms{76.0}{3.5} & \seedms{75.0}{5.3} & \bestseedms{82.0}{1.7} & \bestseedms{83.0}{1.7} & \seedms{76.0}{4.4} & \seeddeltams{+8.0}{3.6} & \seeddeltams{-6.0}{5.3} \\
Place A2B Left & \seedms{76.0}{4.6} & \seedms{80.0}{5.0} & \seedms{79.0}{3.6} & \seedms{77.0}{5.0} & \bestseedms{80.0}{3.6} & \bestseedms{83.0}{2.6} & \seeddeltams{+1.0}{6.9} & \seeddeltams{+6.0}{2.6} \\
Place A2B Right & \seedms{77.0}{4.4} & \seedms{70.0}{3.0} & \seedms{76.0}{5.2} & \seedms{76.0}{4.0} & \bestseedms{81.0}{3.5} & \bestseedms{79.0}{5.2} & \seeddeltams{+5.0}{7.5} & \seeddeltams{+3.0}{2.6} \\
Place Empty Cup & \bestseedms{100.0}{0.0} & \bestseedms{100.0}{0.0} & \bestseedms{100.0}{0.0} & \seedms{99.0}{1.0} & \bestseedms{100.0}{0.0} & \bestseedms{100.0}{0.0} & \seeddeltams{+0.0}{0.0} & \seeddeltams{+1.0}{1.0} \\
Move Can Pot & \seedms{75.0}{4.4} & \seedms{66.0}{4.4} & \bestseedms{82.0}{3.6} & \bestseedms{90.0}{2.0} & \seedms{75.0}{6.0} & \seedms{75.0}{5.0} & \seeddeltams{-7.0}{4.4} & \seeddeltams{-15.0}{7.0} \\
Place Container Plate & \seedms{96.0}{2.6} & \bestseedms{98.0}{1.0} & \seedms{95.0}{4.6} & \seedms{96.0}{1.7} & \bestseedms{100.0}{0.0} & \bestseedms{98.0}{2.0} & \seeddeltams{+5.0}{4.6} & \seeddeltams{+2.0}{3.6} \\
Press Stapler & \seedms{90.0}{3.5} & \bestseedms{95.0}{5.0} & \bestseedms{92.0}{3.6} & \seedms{94.0}{1.7} & \bestseedms{92.0}{5.6} & \bestseedms{95.0}{4.4} & \seeddeltams{+0.0}{2.0} & \seeddeltams{+1.0}{5.6} \\
Place Phone Stand & \seedms{84.0}{5.6} & \bestseedms{96.0}{1.0} & \bestseedms{91.0}{2.6} & \seedms{91.0}{3.0} & \bestseedms{91.0}{1.7} & \bestseedms{96.0}{2.6} & \seeddeltams{+0.0}{2.6} & \seeddeltams{+5.0}{3.6} \\
Place Fan & \seedms{94.0}{4.0} & \seedms{89.0}{4.6} & \seedms{93.0}{2.6} & \seedms{86.0}{5.6} & \bestseedms{97.0}{3.6} & \bestseedms{91.0}{3.0} & \seeddeltams{+4.0}{1.0} & \seeddeltams{+5.0}{5.3} \\
Rotate QRcode & \seedms{78.0}{5.3} & \seedms{86.0}{2.0} & \seedms{83.0}{6.0} & \bestseedms{87.0}{2.6} & \bestseedms{88.0}{1.0} & \bestseedms{87.0}{3.6} & \seeddeltams{+5.0}{6.6} & \seeddeltams{+0.0}{5.3} \\
Place Object Stand & \seedms{96.0}{3.5} & \seedms{90.0}{3.6} & \seedms{96.0}{1.0} & \seedms{85.0}{2.0} & \bestseedms{98.0}{3.5} & \bestseedms{96.0}{1.7} & \seeddeltams{+2.0}{2.6} & \seeddeltams{+11.0}{2.6} \\
Shake Bottle & \bestseedms{100.0}{0.0} & \seedms{99.0}{1.0} & \seedms{98.0}{1.7} & \seedms{99.0}{1.0} & \bestseedms{100.0}{0.0} & \bestseedms{100.0}{0.0} & \seeddeltams{+2.0}{1.7} & \seeddeltams{+1.0}{1.0} \\
Scan Object & \bestseedms{57.0}{5.6} & \seedms{47.0}{3.6} & \seedms{56.0}{4.6} & \seedms{38.0}{2.6} & \seedms{49.0}{2.0} & \bestseedms{57.0}{4.4} & \seeddeltams{-7.0}{4.4} & \seeddeltams{+19.0}{3.0} \\
Pick Diverse Bottles & \seedms{78.0}{5.3} & \seedms{69.0}{2.0} & \seedms{80.0}{3.6} & \seedms{74.0}{5.0} & \bestseedms{83.0}{3.5} & \bestseedms{80.0}{6.0} & \seeddeltams{+3.0}{1.0} & \seeddeltams{+6.0}{11.0} \\
Place Bread Skillet & \bestseedms{91.0}{1.0} & \bestseedms{91.0}{6.0} & \seedms{89.0}{1.0} & \bestseedms{91.0}{2.6} & \seedms{89.0}{6.0} & \seedms{87.0}{5.3} & \seeddeltams{+0.0}{6.6} & \seeddeltams{-4.0}{4.6} \\
Place Bread Basket & \seedms{90.0}{4.6} & \seedms{90.0}{3.6} & \seedms{91.0}{3.6} & \bestseedms{91.0}{5.3} & \bestseedms{92.0}{5.0} & \bestseedms{91.0}{4.6} & \seeddeltams{+1.0}{6.9} & \seeddeltams{+0.0}{7.0} \\
Place Burger Fries & \seedms{96.0}{2.0} & \seedms{96.0}{3.6} & \seedms{95.0}{2.0} & \bestseedms{99.0}{1.0} & \bestseedms{99.0}{1.0} & \seedms{97.0}{5.2} & \seeddeltams{+4.0}{2.6} & \seeddeltams{-2.0}{5.3} \\
Place Cans Plasticbox & \seedms{98.0}{2.6} & \seedms{98.0}{2.6} & \bestseedms{99.0}{1.0} & \seedms{96.0}{4.0} & \bestseedms{99.0}{1.0} & \bestseedms{99.0}{1.0} & \seeddeltams{+0.0}{1.0} & \seeddeltams{+3.0}{5.0} \\
Put Bottles Dustbin & \seedms{80.0}{4.4} & \seedms{83.0}{4.0} & \seedms{88.0}{2.6} & \seedms{87.0}{3.5} & \bestseedms{94.0}{1.0} & \bestseedms{93.0}{5.0} & \seeddeltams{+6.0}{3.0} & \seeddeltams{+6.0}{6.1} \\
Hanging Mug & \seedms{72.0}{2.6} & \seedms{76.0}{5.3} & \seedms{75.0}{2.0} & \bestseedms{82.0}{5.3} & \bestseedms{83.0}{4.6} & \seedms{76.0}{2.0} & \seeddeltams{+8.0}{4.4} & \seeddeltams{-6.0}{6.9} \\
\midrule
\textbf{Average} & \seedms{84.6}{0.6} & \seedms{83.6}{0.4} & \seedms{86.4}{0.7} & \seedms{85.7}{0.3} & \bestseedms{88.8}{0.2} & \bestseedms{87.9}{0.5} & \seeddeltams{+2.4}{0.4} & \seeddeltams{+2.2}{0.2} \\
\bottomrule
\end{tabular}
}
\end{table}

\clearpage
\revadd{\section{Real-Robot Rollout Uncertainty}}
\label[appendix]{app:realrobot_ci}

\revadd{For the real-robot study in \cref{sec:real_robot}, we quantify rollout sampling
uncertainty using 95\% Wilson binomial confidence intervals.
\cref{tab:real_robot_wilson_ci} reports the interval for every method--task cell, computed
from that cell's 100 rollouts, and for every pooled Overall entry, computed from that
method's 300 rollouts. The pooled intervals are $[36.6, 47.7]$ for EvoScene-VLA and
$[32.1, 42.9]$ for LingBot-VLA$^{*}$.}

\begin{table}[H]
\centering
\caption{\revadd{\textbf{Real-robot rollout uncertainty.} For every method--task pair we report the SR (\%) over 100 physical closed-loop rollouts and its 95\% Wilson binomial confidence interval. The Overall row pools the 300 rollouts of each method (100 per task); the full study comprises 1{,}200 rollouts across the four methods. These intervals quantify rollout sampling uncertainty.}}
\label{tab:real_robot_wilson_ci}
\scriptsize
\setlength{\tabcolsep}{3pt}
\renewcommand{\arraystretch}{1.15}
\resizebox{\textwidth}{!}{%
\begin{tabular}{@{}l cc|cc|cc|cc@{}}
\toprule
& \multicolumn{2}{c|}{$\pi_{0.5}$} & \multicolumn{2}{c|}{LingBot-VLA} & \multicolumn{2}{c|}{LingBot-VLA$^{*}$} & \multicolumn{2}{c}{\textbf{EvoScene-VLA (Ours)}} \\
\cmidrule(lr){2-3}\cmidrule(lr){4-5}\cmidrule(lr){6-7}\cmidrule(lr){8-9}
\textbf{Task} & SR (\%) & 95\% CI & SR (\%) & 95\% CI & SR (\%) & 95\% CI & SR (\%) & 95\% CI \\
\midrule
Mirror & 28.0 & $[20.1,\,37.5]$ & 27.0 & $[19.3,\,36.4]$ & 26.0 & $[18.4,\,35.4]$ & \textbf{29.0} & $[21.0,\,38.5]$ \\
Sink & 42.0 & $[32.8,\,51.8]$ & 44.0 & $[34.7,\,53.8]$ & 49.0 & $[39.4,\,58.7]$ & \textbf{51.0} & $[41.3,\,60.6]$ \\
Cutting Board & 44.0 & $[34.7,\,53.8]$ & 34.0 & $[25.5,\,43.7]$ & 37.0 & $[28.2,\,46.8]$ & \textbf{46.0} & $[36.6,\,55.7]$ \\
\midrule
\textbf{Overall} & 38.0 & $[32.7,\,43.6]$ & 35.0 & $[29.8,\,40.6]$ & 37.3 & $[32.1,\,42.9]$ & \textbf{42.0} & $[36.6,\,47.7]$ \\
\bottomrule
\end{tabular}
}
\end{table}

\FloatBarrier
\revadd{\section{\locnew{Real-Robot Closed-Loop Rollouts}}}
\label[appendix]{app:realrobot_rollouts}

\begin{figure}[H]
\centering
\setlength{\tabcolsep}{1pt}
\setlength{\fboxsep}{0pt}\setlength{\fboxrule}{0.3pt}

\begin{tabular}{@{}*{4}{>{\centering\arraybackslash}m{1.31in}}@{}}
\multicolumn{4}{@{}l@{}}{\small\textbf{Rollout 1 (failure)}} \\[1pt]
\fbox{\includegraphics[width=1.29in]{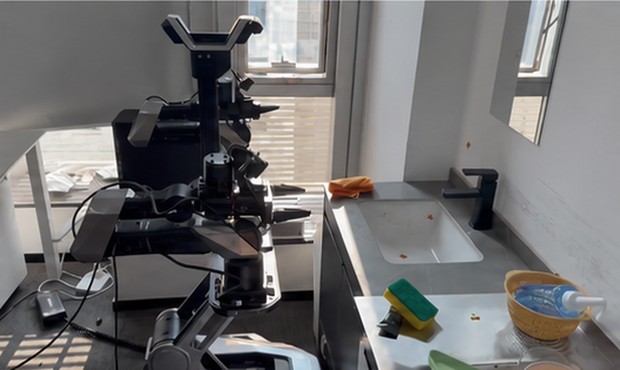}} &
\fbox{\includegraphics[width=1.29in]{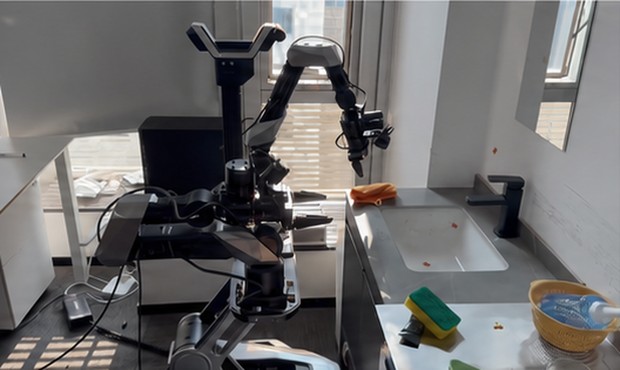}} &
\fbox{\includegraphics[width=1.29in]{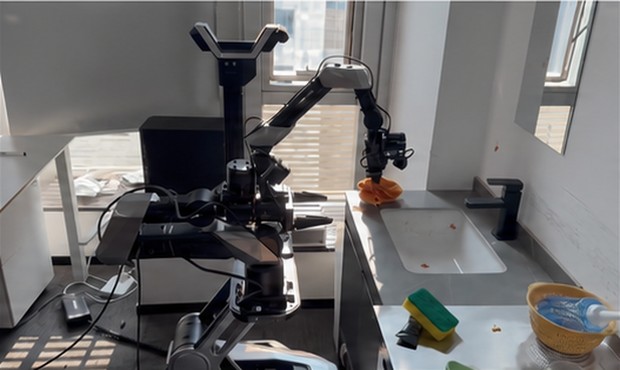}} &
\fbox{\includegraphics[width=1.29in]{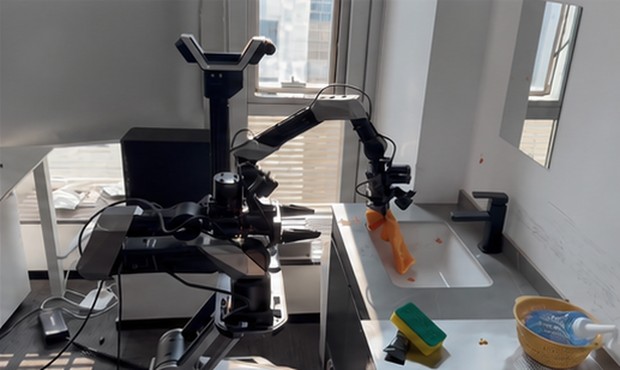}} \\[1pt]
\fbox{\includegraphics[width=1.29in]{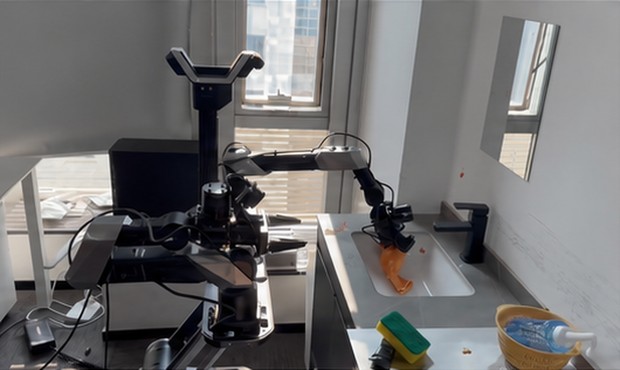}} &
\fbox{\includegraphics[width=1.29in]{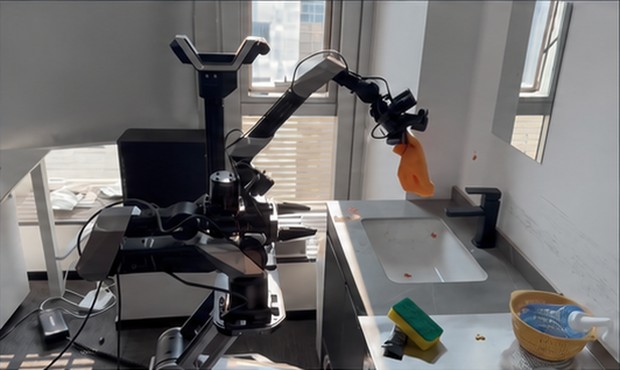}} &
\fbox{\includegraphics[width=1.29in]{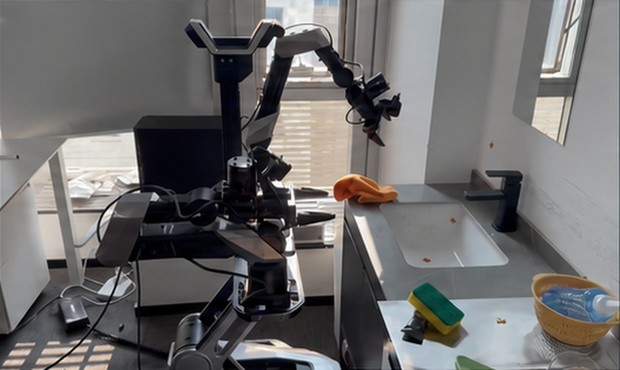}} &
\fbox{\includegraphics[width=1.29in]{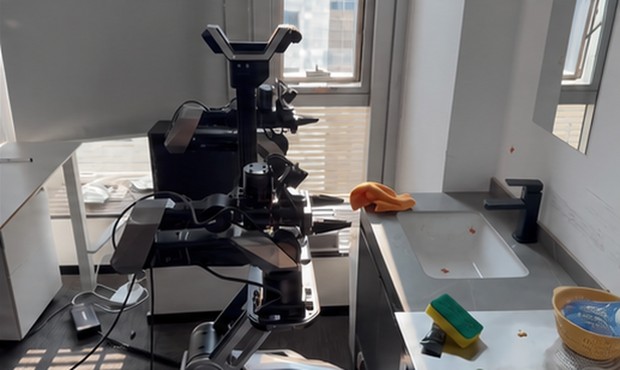}} \\[5pt]
\multicolumn{4}{@{}l@{}}{\small\textbf{Rollout 2 (success)}} \\[1pt]
\fbox{\includegraphics[width=1.29in]{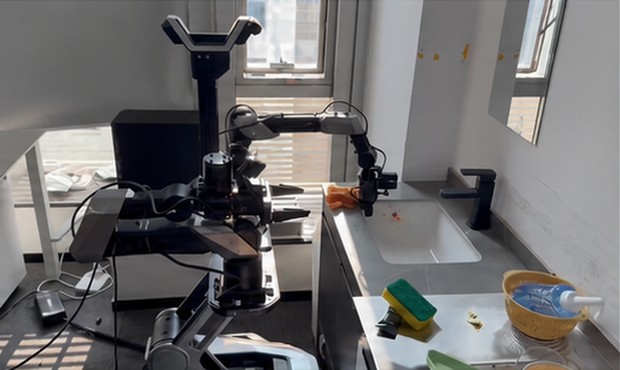}} &
\fbox{\includegraphics[width=1.29in]{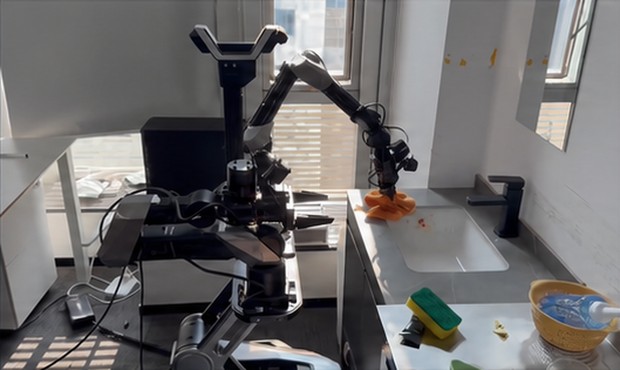}} &
\fbox{\includegraphics[width=1.29in]{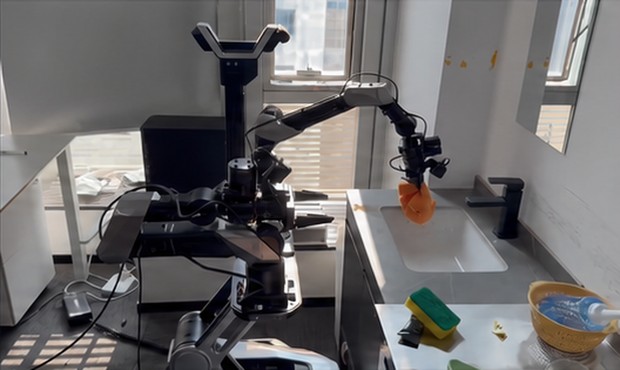}} &
\fbox{\includegraphics[width=1.29in]{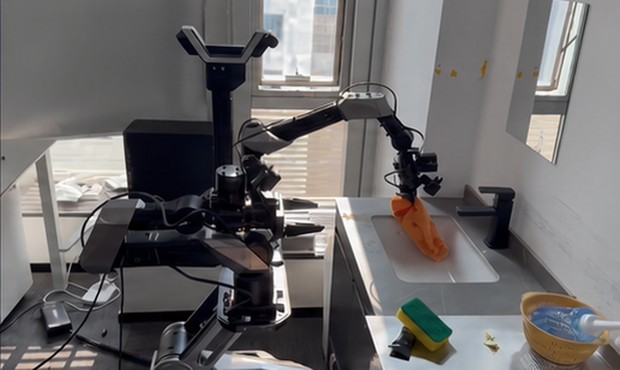}} \\[1pt]
\fbox{\includegraphics[width=1.29in]{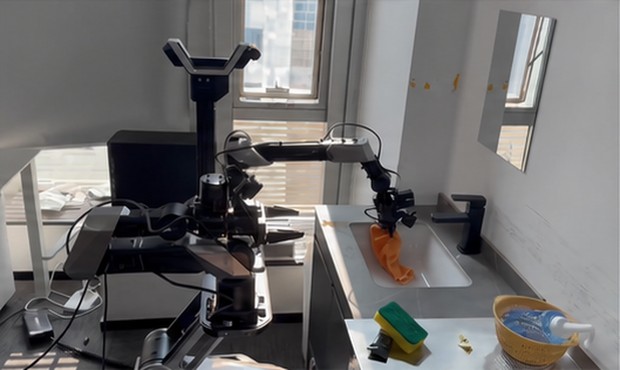}} &
\fbox{\includegraphics[width=1.29in]{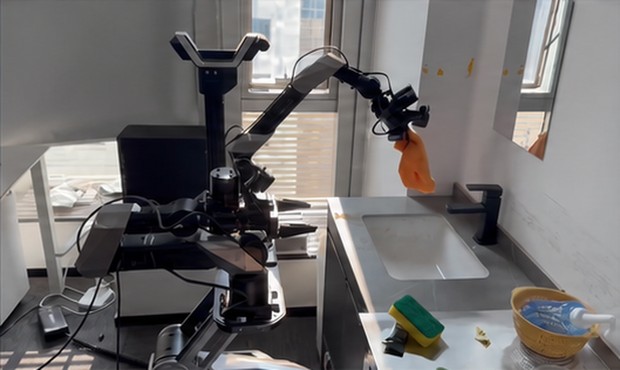}} &
\fbox{\includegraphics[width=1.29in]{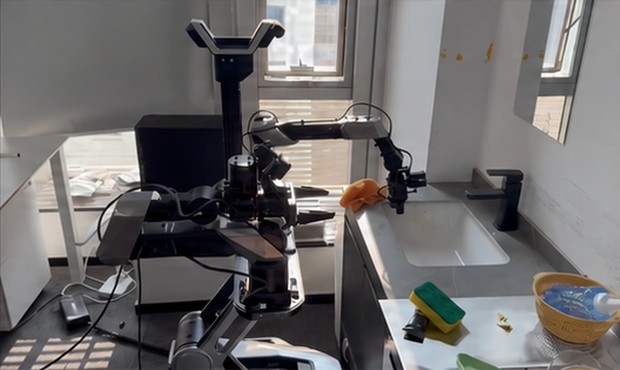}} &
\fbox{\includegraphics[width=1.29in]{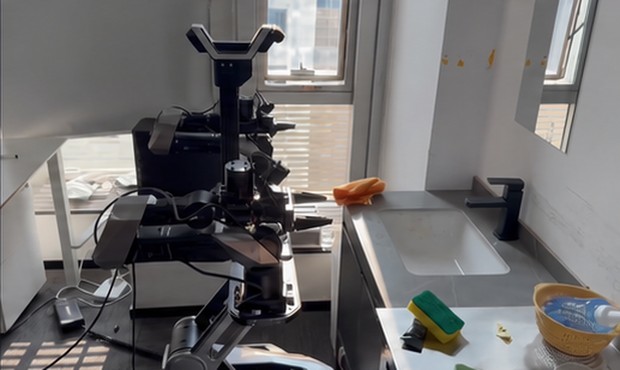}} \\
\end{tabular}

\caption{\locnew{\textbf{Real-robot closed-loop rollouts on \emph{Clean The Sink}.} Two
executions of EvoScene-VLA on Galaxea R1-Lite, read left to right and top to bottom
within each rollout. The robot picks up the cloth, transports it to the basin, and wipes
across the surface. Rollout 1 fails with visible ketchup remaining in
the basin at the end, whereas Rollout 2 clears the basin and succeeds.}}
\label{fig:realrobot_rollouts}
\end{figure}

\locnew{\revadd{\cref{fig:realrobot_rollouts} contrasts a failed
execution (Rollout 1) with the successful execution summarized in
\cref{fig:realrobot_rollout_main} (Rollout 2). Both are closed-loop executions of
EvoScene-VLA on \emph{Clean The Sink}, the longest-horizon of the three real-robot tasks. Frames are
sampled uniformly from the head camera over each execution. The task requires the
robot to pick up the cloth, bring it to the basin, and wipe across the surface; because
wiping changes the surface state, later controls must target regions the tool has not yet
covered, which is exactly the condition the recurrent prior is meant to serve. These are
physical executions on Galaxea R1-Lite under the protocol of \cref{sec:real_robot}, not
dataset replay.}}

\clearpage
\section{Additional Trajectory Visualizations}
\label[appendix]{app:trajectories}

\begin{figure}[H]
\centering

\begin{minipage}[t]{0.49\textwidth}
    \centering
    \includegraphics[width=\linewidth]{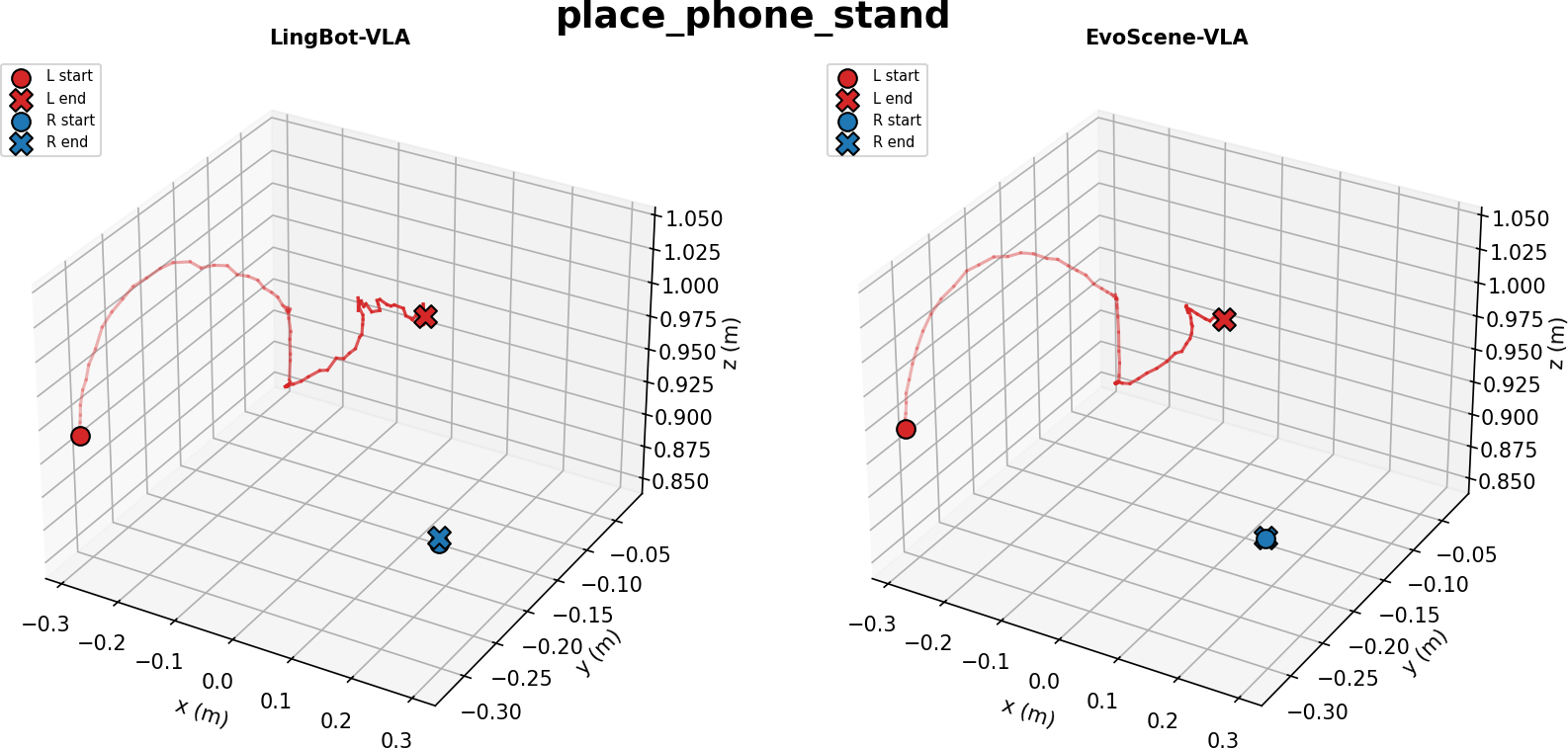}
    \caption*{\small (a) \texttt{place\_phone\_stand}}
\end{minipage}
\hfill
\begin{minipage}[t]{0.49\textwidth}
    \centering
    \includegraphics[width=\linewidth]{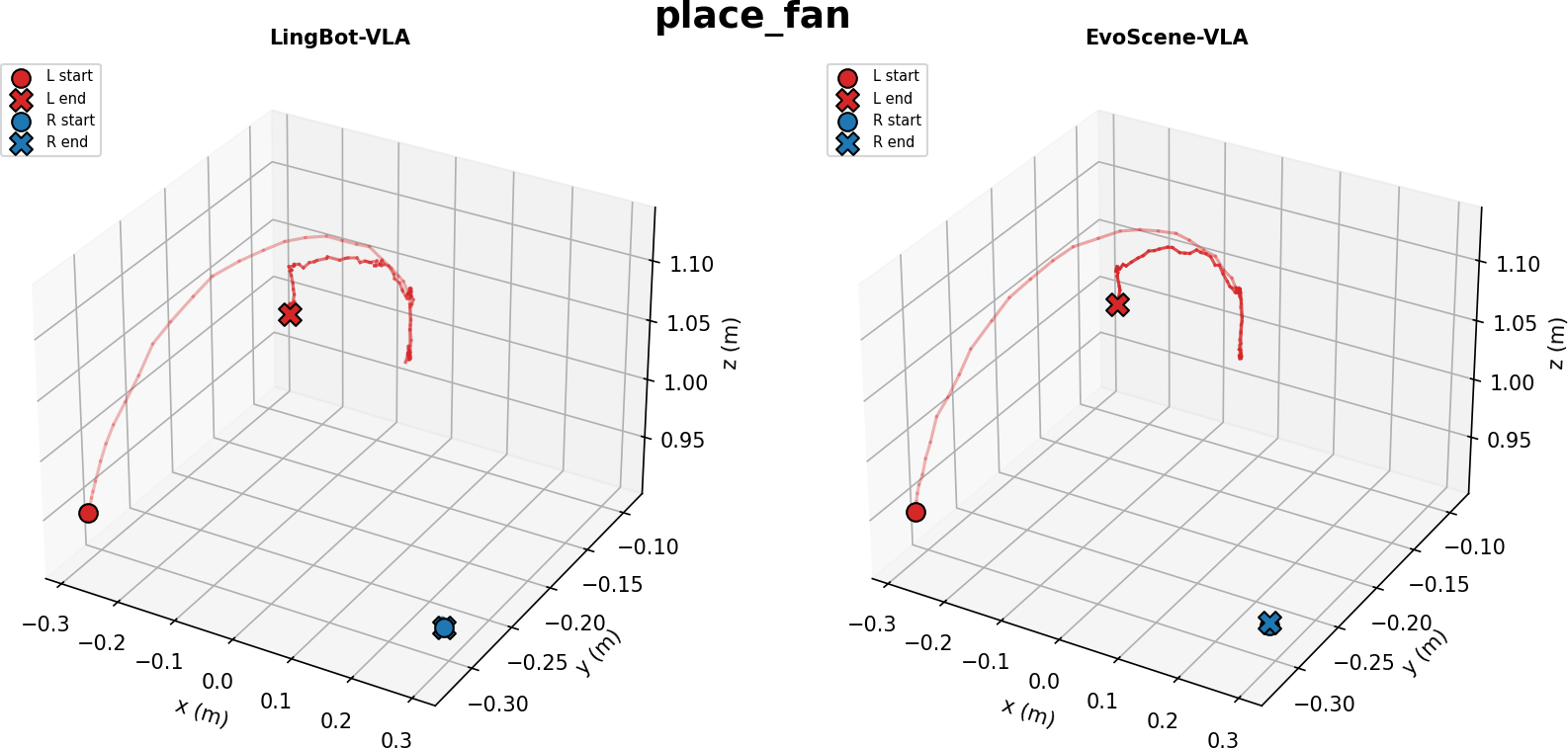}
    \caption*{\small (b) \texttt{place\_fan}}
\end{minipage}

\vspace{2mm}

\begin{minipage}[t]{0.49\textwidth}
    \centering
    \includegraphics[width=\linewidth]{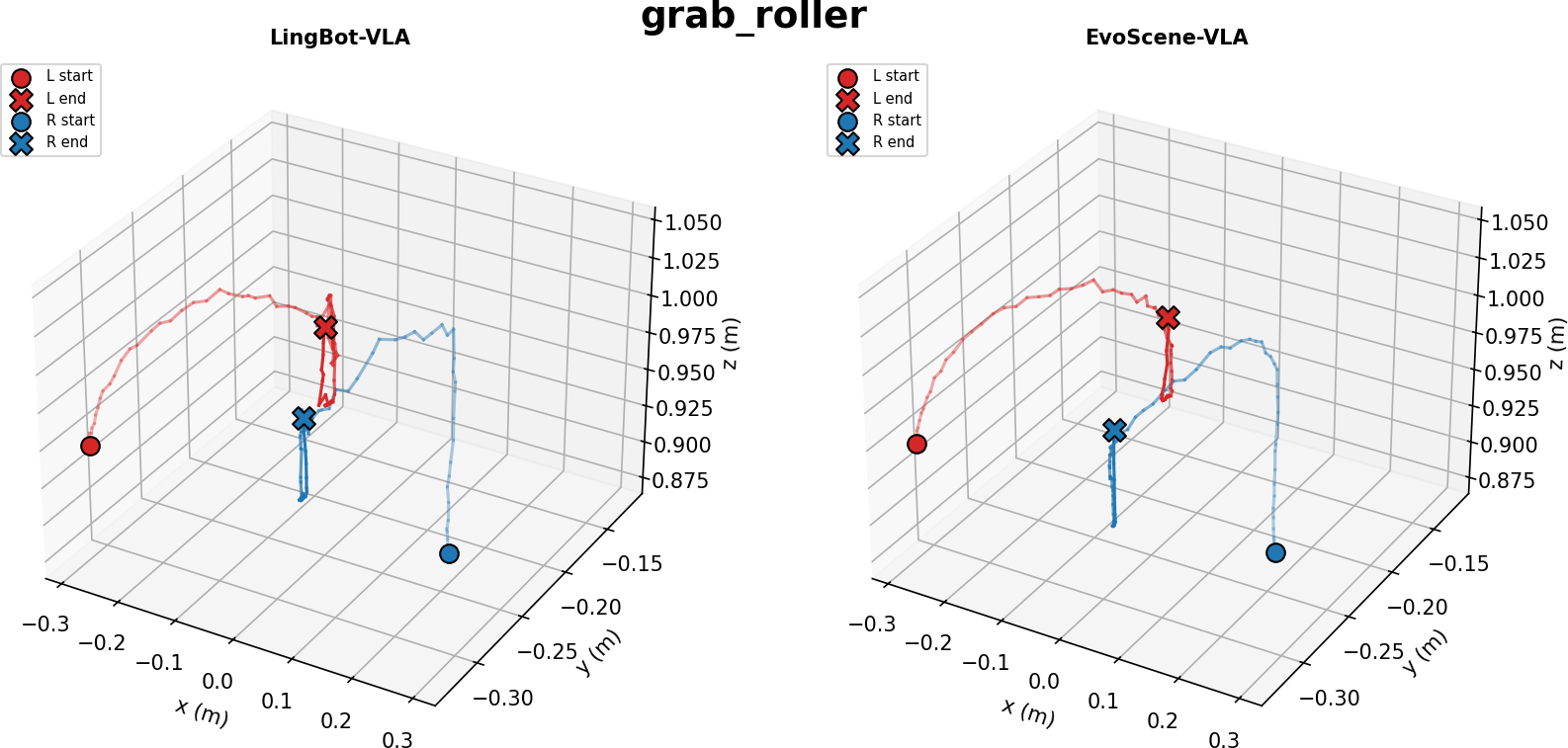}
    \caption*{\small (c) \texttt{grab\_roller}}
\end{minipage}
\hfill
\begin{minipage}[t]{0.49\textwidth}
    \centering
    \includegraphics[width=\linewidth]{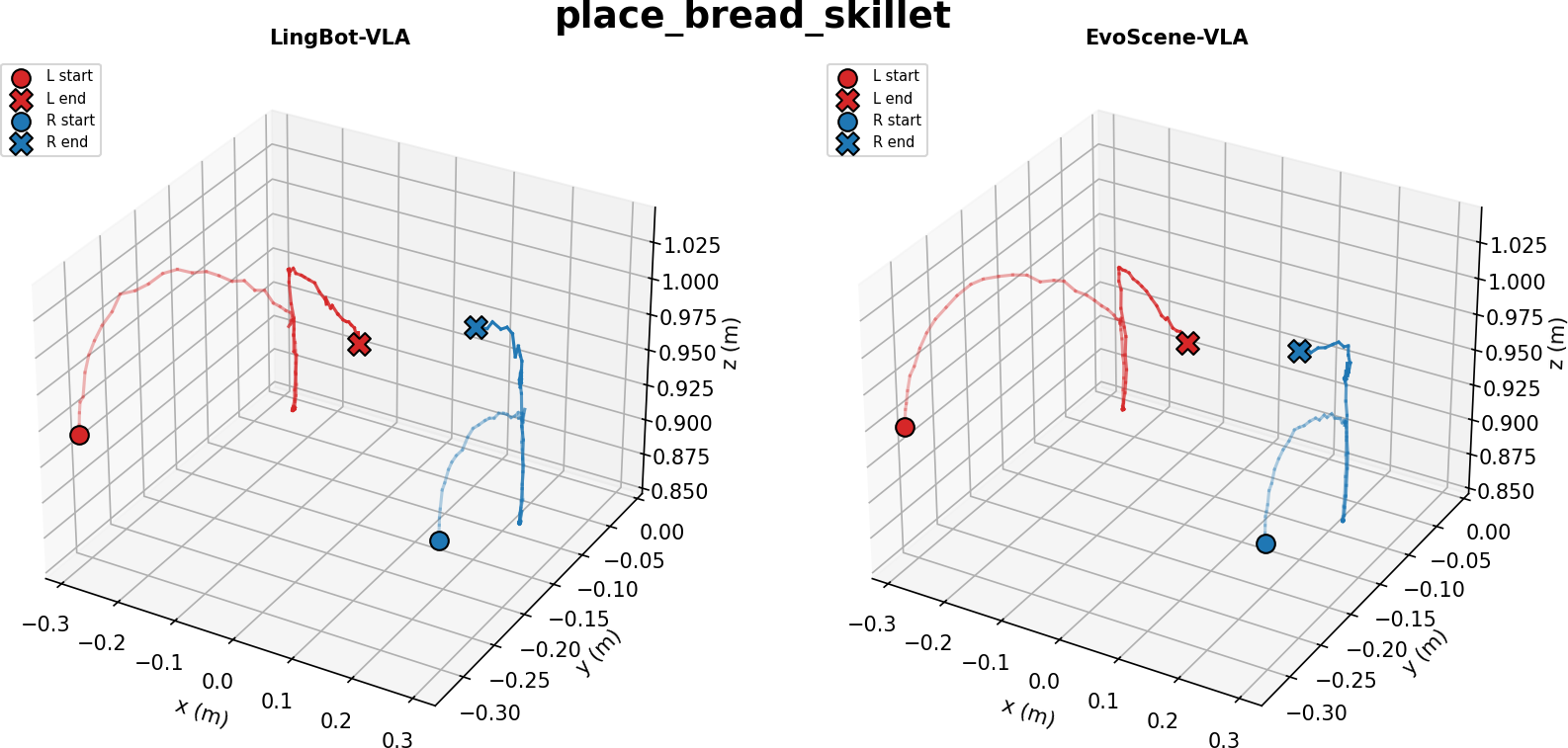}
    \caption*{\small (d) \texttt{place\_bread\_skillet}}
\end{minipage}

\vspace{2mm}

\begin{minipage}[t]{0.49\textwidth}
    \centering
    \includegraphics[width=\linewidth]{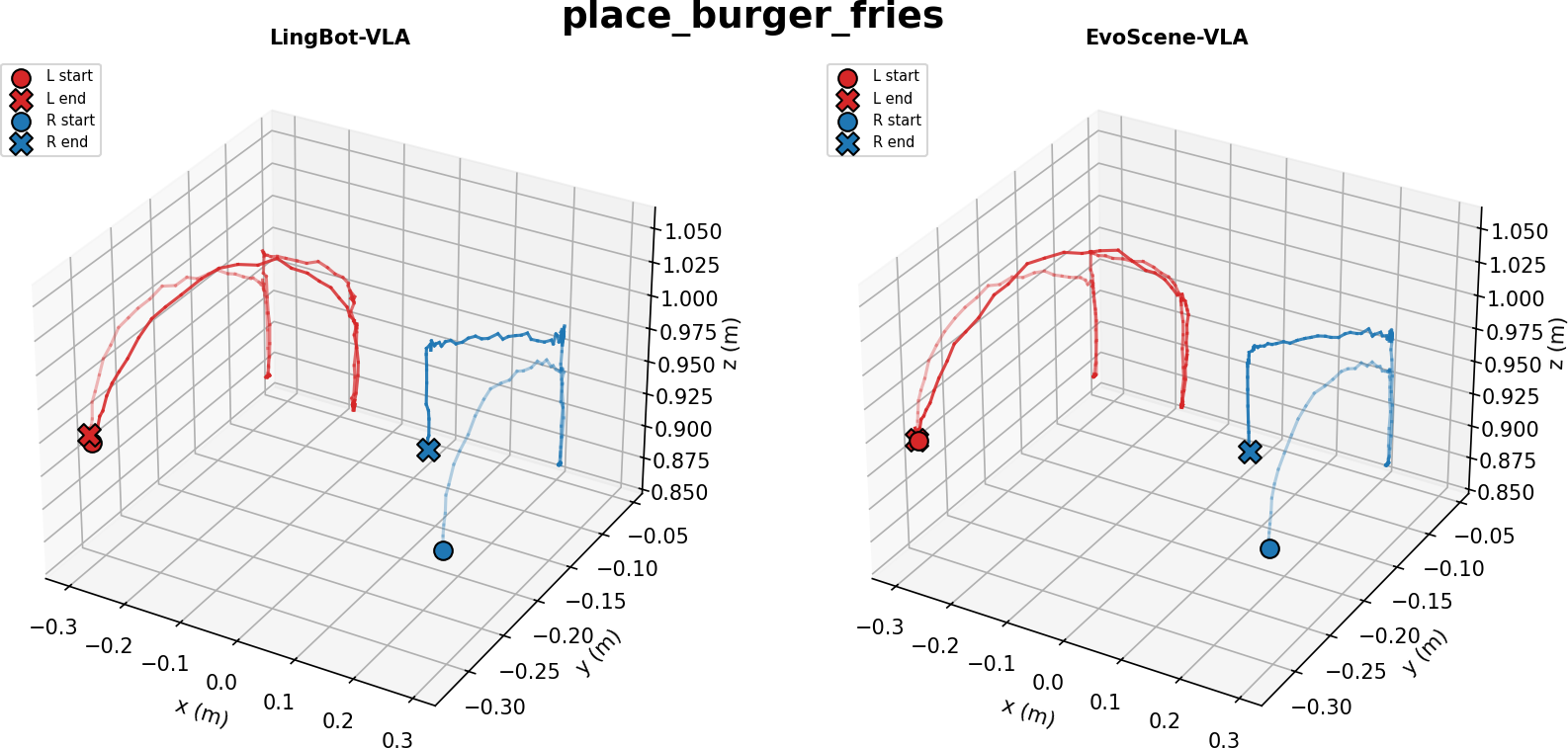}
    \caption*{\small (e) \texttt{place\_burger\_fries}}
\end{minipage}
\hfill
\begin{minipage}[t]{0.49\textwidth}
    \centering
    \includegraphics[width=\linewidth]{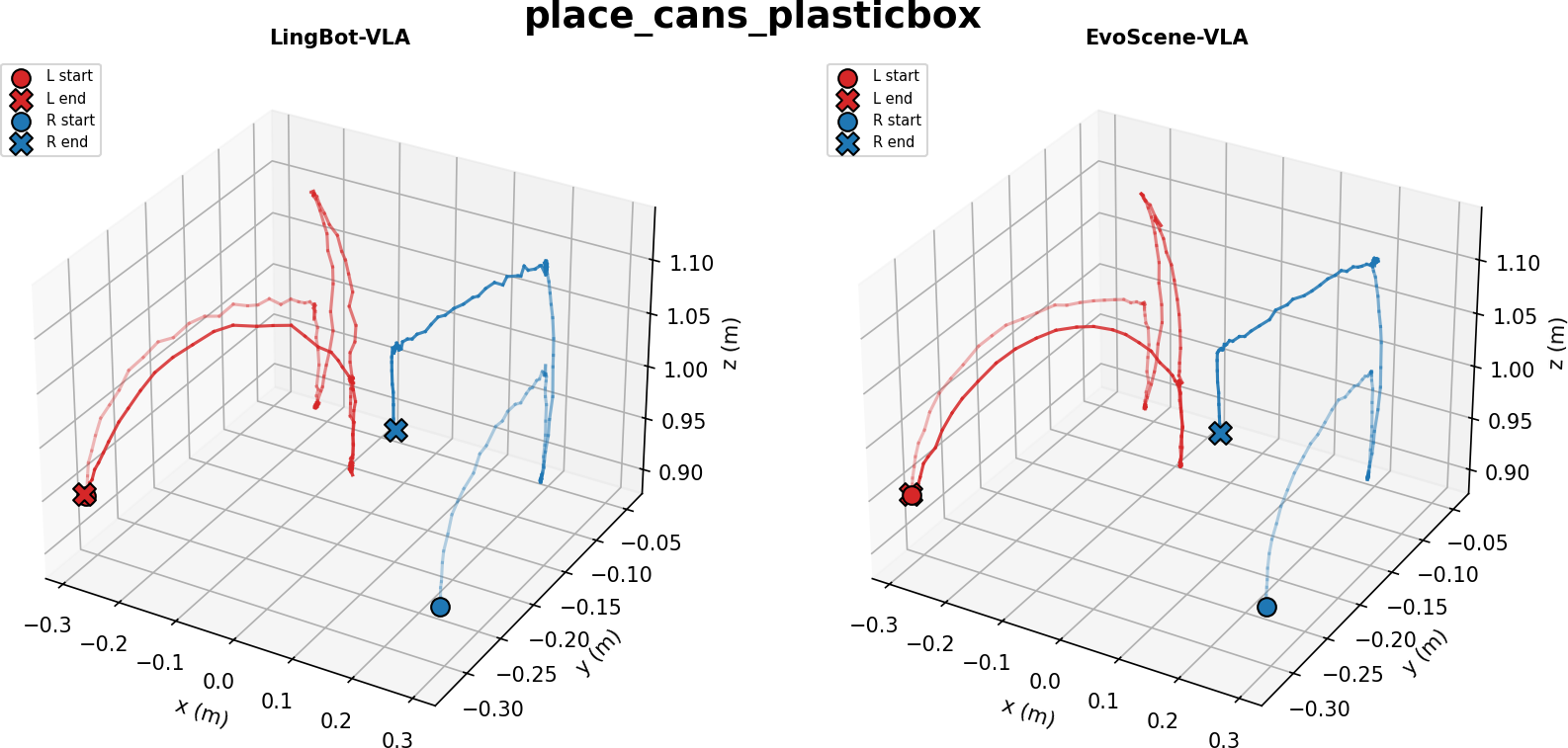}
    \caption*{\small (f) \texttt{place\_cans\_plasticbox}}
\end{minipage}

\vspace{-1mm}
\caption{Additional 3D end-effector trajectory comparisons on six RoboTwin tasks, complementing \cref{fig:trajectory_plots}. EvoScene-VLA's paths appear smoother than LingBot-VLA's in these six qualitative examples.}
\label{fig:appendix_trajectories}
\end{figure}

\cref{fig:appendix_trajectories} shows six additional 3D end-effector trajectory comparisons complementing \cref{fig:trajectory_plots}.

\FloatBarrier
\revadd{\section{\locnew{Decoding Geometry from the Scene Representation on RoboTwin}}}
\label[appendix]{app:robotwin_scene_decode}

\begin{figure}[p]
\centering
\setlength{\tabcolsep}{1.5pt}
\setlength{\fboxsep}{0pt}\setlength{\fboxrule}{0.3pt}
\newcommand{\rlbl}[1]{\begin{minipage}[c]{0.60in}\centering\scriptsize #1\end{minipage}}

\begin{tabular}{@{}>{\centering\arraybackslash}m{0.60in}@{\hspace{2pt}}*{3}{>{\centering\arraybackslash}m{1.60in}}@{}}
 & \small\textbf{RGB} & \small\textbf{3DFM reference} & \small\textbf{Decoded from $s_p$} \\[2pt]

\rlbl{\texttt{place\_cans\_}\\\texttt{plasticbox}\\[2pt]High} &
\fbox{\includegraphics[width=1.58in]{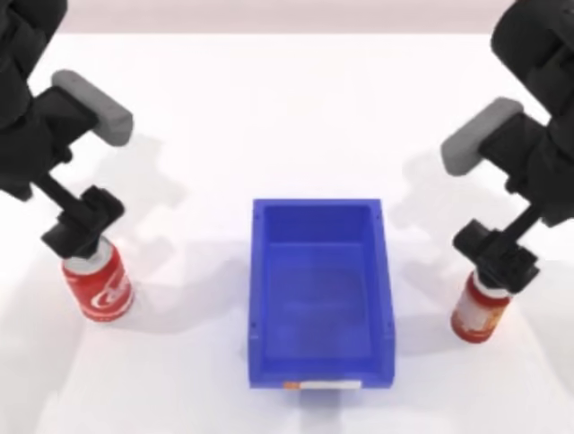}} &
\fbox{\includegraphics[width=1.58in]{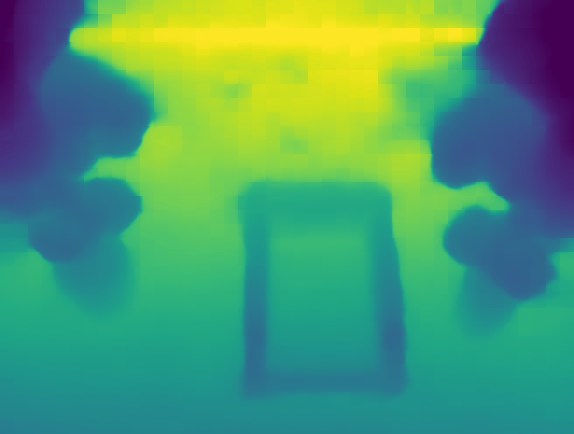}} &
\fbox{\includegraphics[width=1.58in]{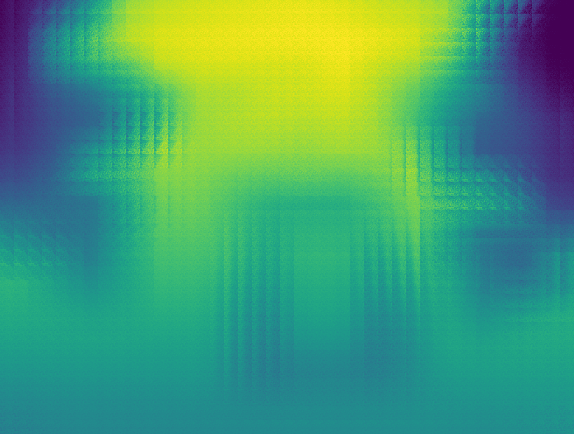}} \\[2pt]

\rlbl{Left wrist} &
\fbox{\includegraphics[width=1.58in]{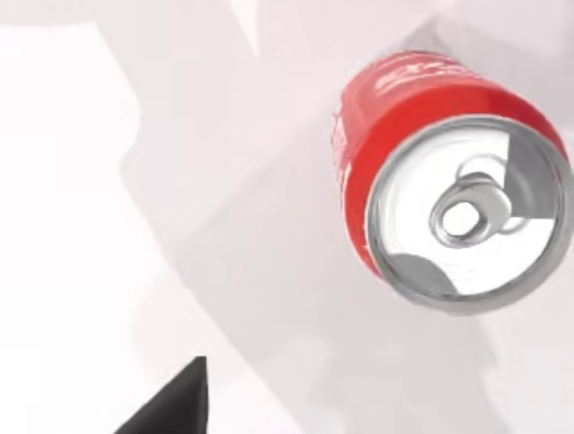}} &
\fbox{\includegraphics[width=1.58in]{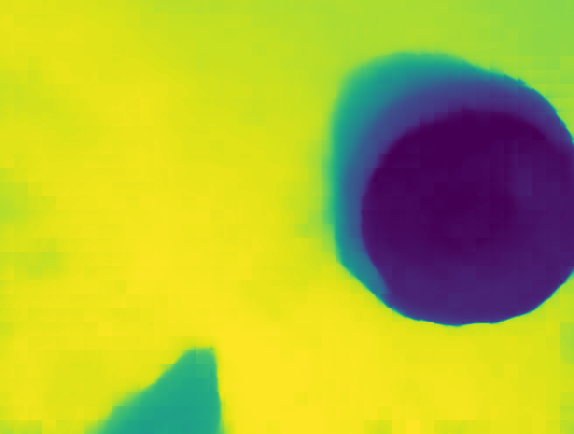}} &
\fbox{\includegraphics[width=1.58in]{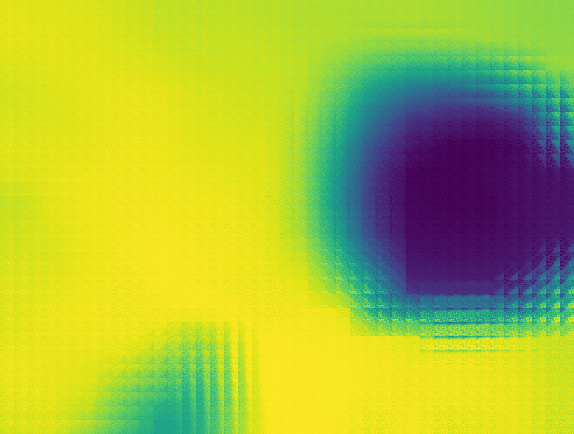}} \\[2pt]

\rlbl{Right wrist} &
\fbox{\includegraphics[width=1.58in]{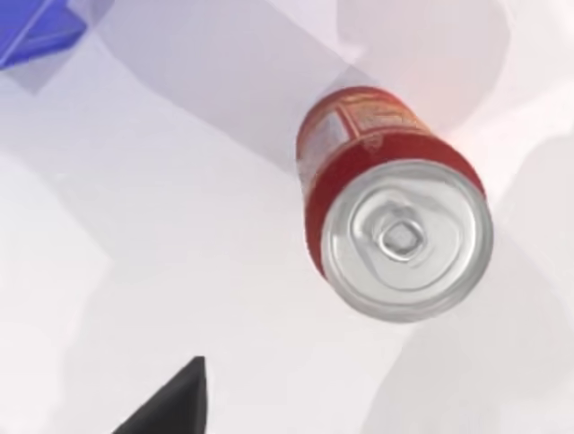}} &
\fbox{\includegraphics[width=1.58in]{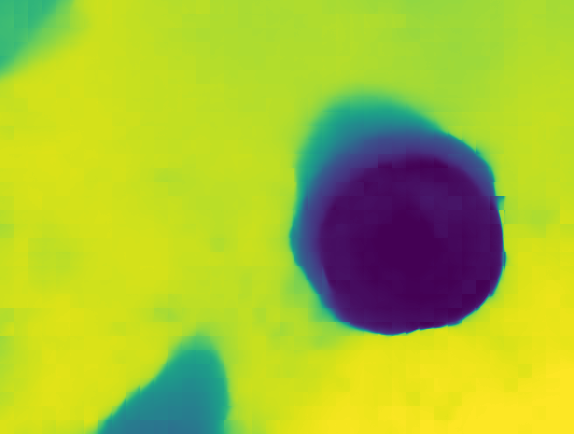}} &
\fbox{\includegraphics[width=1.58in]{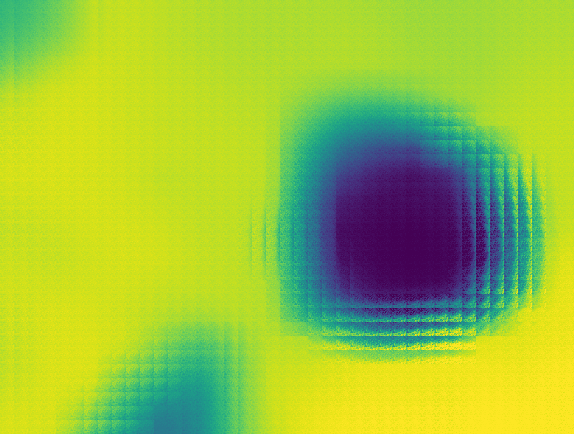}} \\[6pt]

\rlbl{\texttt{click\_bell}\\[2pt]High} &
\fbox{\includegraphics[width=1.58in]{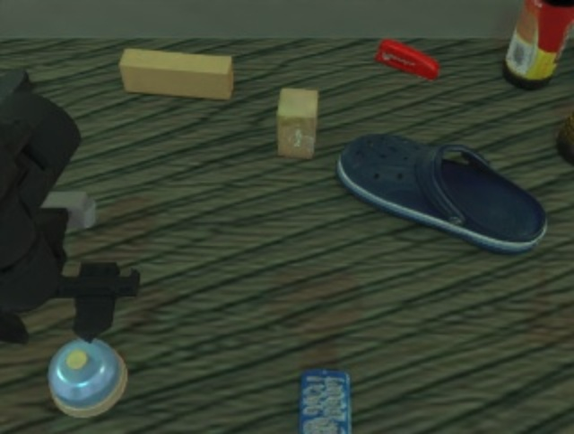}} &
\fbox{\includegraphics[width=1.58in]{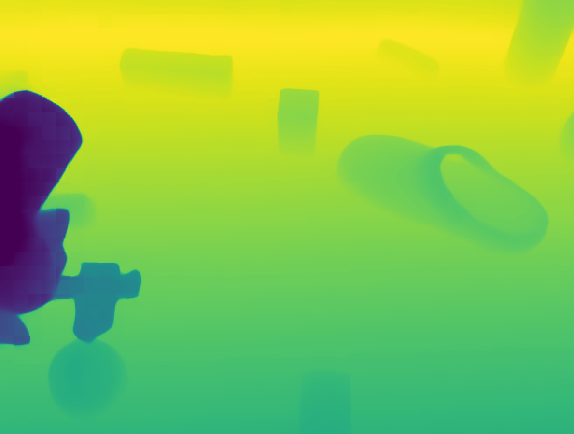}} &
\fbox{\includegraphics[width=1.58in]{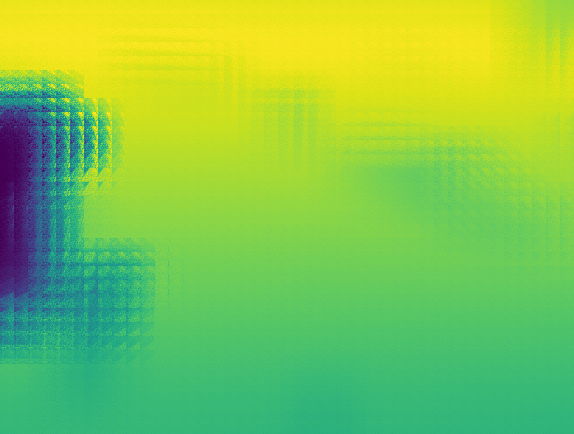}} \\[2pt]

\rlbl{Left wrist} &
\fbox{\includegraphics[width=1.58in]{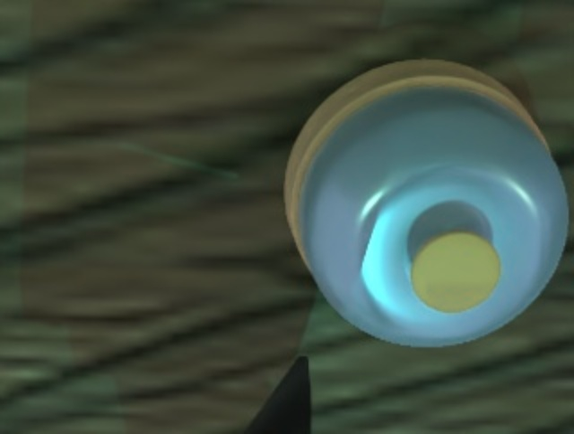}} &
\fbox{\includegraphics[width=1.58in]{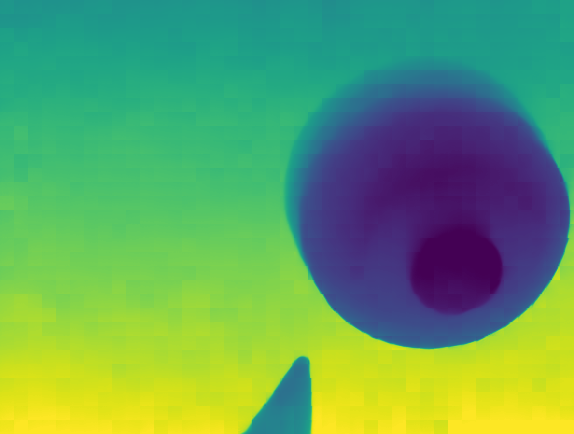}} &
\fbox{\includegraphics[width=1.58in]{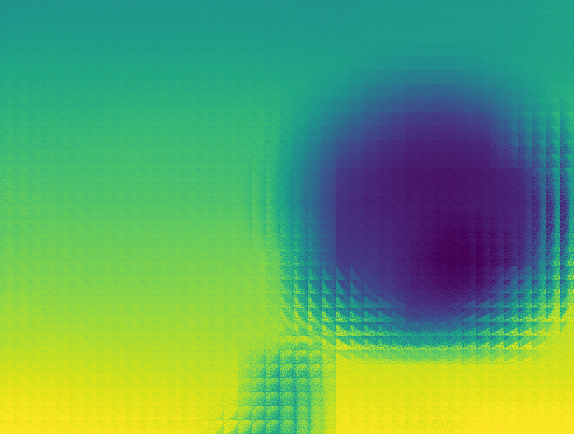}} \\[2pt]

\rlbl{Right wrist} &
\fbox{\includegraphics[width=1.58in]{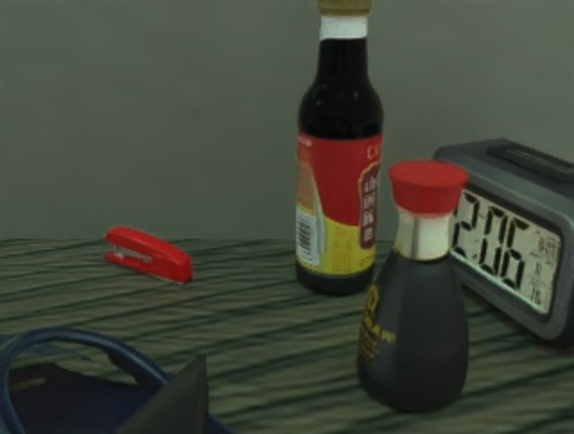}} &
\fbox{\includegraphics[width=1.58in]{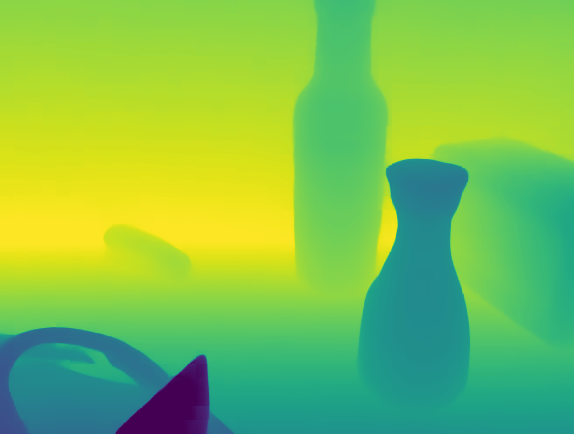}} &
\fbox{\includegraphics[width=1.58in]{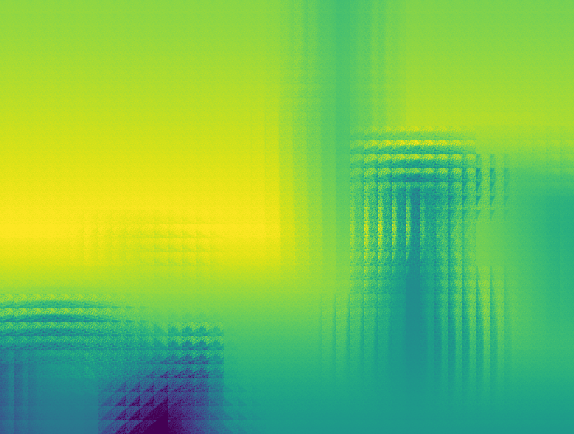}} \\
\end{tabular}

\caption{\locnew{\textbf{Geometry decoded from the recurrent scene representation.}
Two RoboTwin frames with three camera views each. \emph{3DFM reference} decodes
the frozen features $Z_t$, while \emph{Decoded from $s_p$} applies the same
frozen point head to $P_t$ produced from $s_p$. Each pair shares a colour range
fitted on its reference. The maps retain coarse scene layout, support
surfaces, and relative depth ordering, but fine details remain incomplete.}}
\label{fig:robotwin_scene_decode}
\end{figure}

\locnew{\revadd{The global anchor in \cref{sec:globalanchor} trains $P_t$, which the
view-conditioned decoder produces from $s_p$, to match frozen 3DFM features
$Z_t$. To inspect the learned representation without fitting an additional
probe, we pass $P_t$ and $Z_t$ through the same frozen 3DFM point head.
\cref{fig:robotwin_scene_decode} renders log depth across two RoboTwin frames
and all three camera views. Each pair uses a shared colour range fitted on its
3DFM reference, so independent normalization cannot make the two outputs
appear more similar.}}

\clearpage
\revadd{\section{\locnew{Depth Readouts and Prior Interventions}}}
\label[appendix]{app:scene_depth}

\locnew{\revadd{\cref{app:robotwin_scene_decode} shows qualitative RoboTwin decodes. This
section quantifies the same question on LIBERO. Because $P_t$ is trained to match $Z_t$, it lies in the frozen
3DFM's feature space, so its point head decodes it to depth
without fitting to our data. A favourable score therefore cannot be
attributed to a probe that learned depth on its own. The reference depth is the
same head applied to $Z_t$.}}

\paragraph{\locnew{\revadd{Protocol.}}}
\locnew{\revadd{We evaluate all $277{,}713$ frames from $1{,}712$ episodes
in the available four-suite LIBERO no-noops training corpus, giving $555{,}426$
images over two views per condition. These are training-set readout diagnostics.
Each episode is evaluated using independent offset streams to cover every frame.
Each stream starts with the learned initial prior and resets at the episode
boundary. The cold condition resets the prior at every frame.
Both $P_t$ and $Z_t$ are layer-normalized
before their comparison in $\mathcal{L}_{\mathrm{rep}}$ (\cref{eq:l_rep}), so
this loss compares standardized current-frame features rather than their raw
norms. $\mathcal{L}_{\mathrm{pred}}$ compares unnormalized future-frame
features through the shared decoder (\cref{eq:l_pred}), so the training
objective also contains scale-sensitive supervision.
For AbsRel, $\delta_1$, and $\delta_2$, we align each model prediction to the
reference with a least-squares scale and shift in log depth before scoring.
Pearson is the signed correlation of raw predicted and reference log depth,
before alignment. The constant control predicts each image's median reference
depth without an additional affine fit; its Pearson correlation is undefined.
We average per-image scores within each episode, then average episodes equally.
These readouts quantify relative geometry with respect to the frozen 3DFM reference.}}

The depth-readout metrics are reported in \cref{tab:scene_depth} in the main paper.

\begin{figure}[p]
\centering
\setlength{\tabcolsep}{1.5pt}
\setlength{\fboxsep}{0pt}\setlength{\fboxrule}{0.3pt}
\newcommand{\vlbl}[1]{\scriptsize #1}
\begin{tabular}{@{}>{\centering\arraybackslash}m{0.62in}@{\hspace{2pt}}*{3}{>{\centering\arraybackslash}m{1.58in}}@{}}
 & \small\textbf{RGB} & \small\textbf{Reference} & \small\textbf{Decoded from $s_p$} \\[2pt]
\vlbl{Agent view} &
\fbox{\includegraphics[width=1.56in]{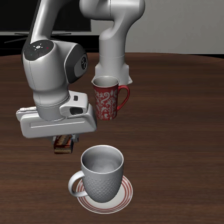}} &
\fbox{\includegraphics[width=1.56in]{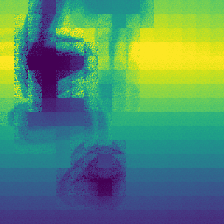}} &
\fbox{\includegraphics[width=1.56in]{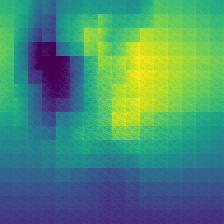}} \\[2pt]
\vlbl{Agent view} &
\fbox{\includegraphics[width=1.56in]{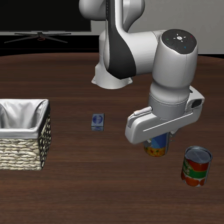}} &
\fbox{\includegraphics[width=1.56in]{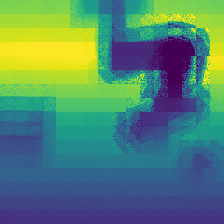}} &
\fbox{\includegraphics[width=1.56in]{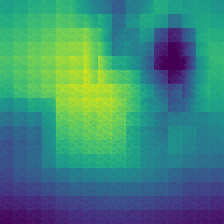}} \\[2pt]
\vlbl{Wrist view} &
\fbox{\includegraphics[width=1.56in]{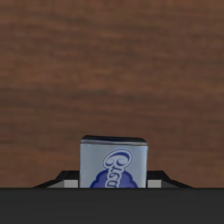}} &
\fbox{\includegraphics[width=1.56in]{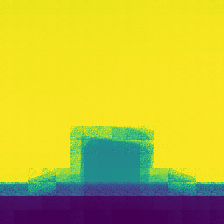}} &
\fbox{\includegraphics[width=1.56in]{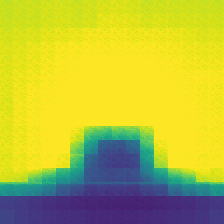}} \\[2pt]
\vlbl{Wrist view} &
\fbox{\includegraphics[width=1.56in]{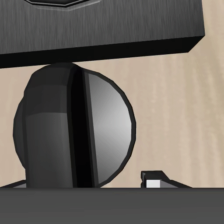}} &
\fbox{\includegraphics[width=1.56in]{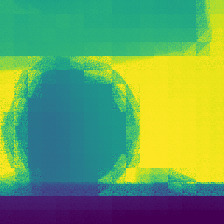}} &
\fbox{\includegraphics[width=1.56in]{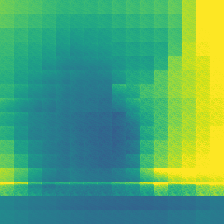}} \\
\end{tabular}
\caption{\locnew{\textbf{Depth decoded from the scene representation.} Four LIBERO
images covering two agent views and two wrist views. \emph{Reference} decodes the frozen 3DFM target;
\emph{Decoded} decodes $P_t$ through the same frozen head. Each row shares a
colour range fitted on its reference.}}
\label{fig:scene_depth}
\end{figure}

\locnew{\revadd{The representation carries geometry: $\delta_1$ is $0.732$ against $0.586$ for
the constant baseline, and AbsRel is $0.163$ against $0.411$.
\cref{fig:scene_depth} shows the corresponding decodes. Coarse layout, support
surfaces, and arm placement survive; object-level detail does not, and the
$16 \times 16$ token grid leaves visible block structure. Wrist views are
consistently weaker than the agent view.}}

\paragraph{\locnew{\revadd{What the recurrent prior contributes.}}}
\locnew{\revadd{\cref{tab:scene_depth} shows a gain from carrying the
prior across chunks with both original camera images as input: $\delta_1$ rises from $0.701$
with a cold prior to $0.732$ with the recurrent prior, a gain of $3.1$
percentage points. $\delta_2$ also rises from $0.919$ to $0.930$, while
AbsRel falls from $0.172$ to $0.163$.
This comparison tests carrying the prior against resetting it to the first-chunk
initialization. We next test whether the content of the incoming prior matters
to closed-loop control.}}

\paragraph{\locnew{\revadd{Closed-loop prior interventions.}}}
\locnew{\revadd{\cref{tab:prior_content} compares task success under
\emph{matched} and \emph{mismatched} incoming priors. The matched condition uses
the action-updated scene state carried from the preceding chunk of the current
episode. The mismatched condition replaces this state with a clean prior
sampled from another episode, allowing donors from other tasks. All conditions
share a clean initial action chunk; interventions apply at each subsequent
replan.}}

\locnew{\revadd{We evaluate three camera-input conditions.
\emph{Both views unmodified} uses the current rollout's agent and wrist images.
\emph{Agent view replaced} and \emph{wrist view replaced} substitute the selected
camera image with the corresponding camera image from a randomly selected
frame of another episode; the other camera remains unchanged. Image and prior
donors are sampled independently. Matched and mismatched trials share initial
conditions and the replacement-image schedule, while their actions and later
observations may diverge during closed-loop execution.}}

\locnew{\revadd{We report the percentage of evaluation trials that
complete the task as success rate (SR). $\Delta$ is matched minus mismatched SR
in percentage points. With both views unmodified, SR is $85.45\%$ versus
$83.75\%$, a margin of $1.7$ points. Replacing the agent view yields $14.45\%$
versus $1.65\%$ ($+12.8$ points); replacing the wrist view yields $12.90\%$
versus $2.70\%$ ($+10.2$ points). Matched priors yield higher success in all
three conditions, with larger margins when one view is replaced.}}

\end{document}